\documentclass[twocolumn,letterpaper]{IEEEAerospaceCLS}  

\usepackage{subfigure}
\usepackage[]{graphicx}    
\usepackage{algorithm}
\usepackage{relsize}
\usepackage{algpseudocode}
\usepackage{amsmath}
\usepackage{multirow}
\newcommand{\ignore}[1]{}  
\algnewcommand{\LineComment}[1]{\State \(\triangleright\) #1}
\DeclareMathOperator*{\argmax}{argmax}
\usepackage[final]{pdfpages}

\newcommand{\squeezeup}{\vspace{-2.5mm}}
\begin{document}

\title{Developing Modular Autonomous Capabilities \\for sUAS Operations\thanks{DISTRIBUTION STATEMENT A. Approved for public release. Distribution is unlimited. This material is based upon work supported by the Department of the Air Force under Air Force Contract No. FA8702-15-D-0001. Any opinions, findings, conclusions or recommendations expressed in this material are those of the author(s) and do not necessarily reflect the views of the Department of the Air Force. © 2023 Massachusetts Institute of Technology. Delivered to the U.S. Government with Unlimited Rights, as defined in DFARS Part 252.227-7013 or 7014 (Feb 2014). Notwithstanding any copyright notice, U.S. Government rights in this work are defined by DFARS 252.227-7013 or DFARS 252.227-7014 as detailed above. Use of this work other than as specifically authorized by the U.S. Government may violate any copyrights that exist in this work. \footnotesize 978-1-6654-9032-0/23/$\$31.00$ \copyright2023 IEEE}}

\author{
Keegan Quigley, Virginia Goodwin, Luis Alvarez, Justin Yao \\ 
MIT Lincoln Laboratory\\
244 Wood St\\
Lexington, MA 02421\\
\{keegan.quigley, virginia.goodwin, \\luis.alvarez, justin.yao\}@ll.mit.edu \and
Yousef Salaman Maclara\\
Oregon State University\\
101 Covell Hall \\1691 SW Campus Way \\ Corvallis, OR 97331 \\
salamany@oregonstate.edu}

\maketitle

\thispagestyle{plain}
\pagestyle{plain}




\begin{abstract}
Small teams in the field can benefit from the capabilities provided by small Uncrewed Aerial Systems (sUAS) for missions such as reconnaissance, hostile attribution, remote emplacement, and search and rescue. The mobility, communications, and flexible payload capacity of sUAS can offer teams new levels of situational awareness and enable more highly coordinated missions than previously possible. However, piloting such aircraft for specific missions draws personnel away from other mission-critical tasks, increasing the load on remaining personnel while also increasing complexity of operations. For wider adoption and use of sUAS for security and humanitarian missions, safe and robust autonomy must be employed to reduce this burden on small teams. In this paper, we present the development of the Collaborative-UAS for Hostile Attribution, Surveillance, Emplacement, and Reconnaissance (CHASER) testbed, for rapidly prototyping capabilities that will reduce strain on small teams through sensor-guided autonomous control. We attempt to address autonomy needs that are not filled by commercial sUAS platforms by creating and testing a series of composable modules that can be rapidly reconfigured to support multiple specific missions. Methods implemented and presented here include radar track correlation, on-board computer vision target detection, target position estimation, closed-loop relative position control, and efficient search of a 3D volume for target acquisition. We configure and test a series of these modules in an example mission, executing a fully autonomous chase of an intruding sUAS in live flight, and demonstrating the success of the modularized autonomy approach. We present performance results from simulation or live flight tests for each module. Lastly, we describe the software architecture that we have developed for flexible controls and comment on how the capabilities presented may enable additional missions.
\end{abstract}

\tableofcontents

\section{Introduction}

Over the last decade, significant technological advancements in small Uncrewed Aerial Systems (sUAS) have enabled new commercial and hobbyist applications. However, the commercial sUAS market tends to focus on ease of flight with stable high-resolution imagery, and government missions are vastly overshadowed by the private consumer and commercial market \cite{uasmarket}. While government missions can clearly benefit from the capability this technology provides, the burden of dedicating one person's time exclusively to piloting the sUAS is substantial for small teams in the field. There is strong potential benefit from automating UAS functions such as autonomous navigation in complex environments, at-the-edge data exploitation, automated monitoring/alerting, and flexible payloads. Advances in these technology areas do not overlap strongly with commercial interests and therefore targeted research is needed to enable capabilities for many government missions that would benefit from small UAS.

\subsection{sUAS Autonomy}
While some basic autonomy functions, such as return to launch and following preprogrammed waypoints, have been features of commercial sUAS from early on in development \cite{dronehistory}, commercial sUAS have not been driven by the same mission needs towards robust, reactive autonomous functionality as we see in government applications. In fact the opposite is more true: sUAS were typically purchased by private consumers for recreational flying, and therefore there was not as much market pressure to develop significant reactive autonomous flight capability. This paradigm is rapidly changing, as on-board compute power and modern machine learning algorithms have made sUAS viable tools for industrial applications such as agriculture and infrastructure surveillance \cite{rrinsp}, leading to a growing demand for autonomous functionality \cite{uasmarket}. Still, there is a need for significant development in reactive sUAS autonomy that addresses the complex, niche mission needs of small teams in challenging environments \cite{cpbautonomy,ray2022}. 

The wide difference in capabilities needed for commercial/industrial tasks and tactical missions necessitates the development of dedicated algorithms, architectures, and platforms. DARPA's Fast Lightweight Autonomy (FLA) program first addressed this capability gap, demonstrating novel sUAS capabilities in static, cluttered environments through development of perception, navigation, and mapping algorithms \cite{FLA,darpafla2018}. Like DARPA FLA, we also focus on the algorithms and architecture, utilizing commercially available airframes, processors, and sensors. However, we tackle the challenge of navigation and motion planning in response to a \textit{dynamic} target or objective. For this aim, we develop a set of reactive control and perception modules that have little intersection with commercial capabilities.

\subsection{Operational Use-Cases}
The initial motivation for this effort emerged from an incident at Gatwick Airport in the U.K. from 19-21 December, 2018. A sUAS was observed multiple times incurring on the active runway airspace at the airport, and all flights had to be grounded over the course of two days in the middle of the winter holiday travel season \cite{gatwick}. Given the difficulty that local law enforcement had in apprehending the pilot (the only way to guarantee that the sUAS would stop disrupting safe air travel) we postulated that another sUAS, designed to chase the incurring sUAS and hopefully capture imagery of the pilot, would be useful for air safety and other homeland defense applications, but that it would need to operate autonomously, given how quickly the sUAS may leave the local area/operator line of sight, and the difficulty in a human operator reacting in time and maintaining track on another sUAS.

Beyond autonomous chase, there exist a wide variety of autonomous sUAS behaviors that could benefit various government and law enforcement entities, including autonomous surveillance for search and rescue \cite{dhsdrones}, trace gas or other unique signature detection, communications provision in a signal-denied environment \cite{npstc}, and remote emplacement of objects in hard-to-reach locations. Given the wide variety of potential use cases, including multiple different sensor modalities, it was apparent that a modular, composable approach to system design would be extremely beneficial to rapid development and iteration of autonomous capabilities.

\begin{figure}[h]
	\centering
 	{
 		\includegraphics[width=3.2in]{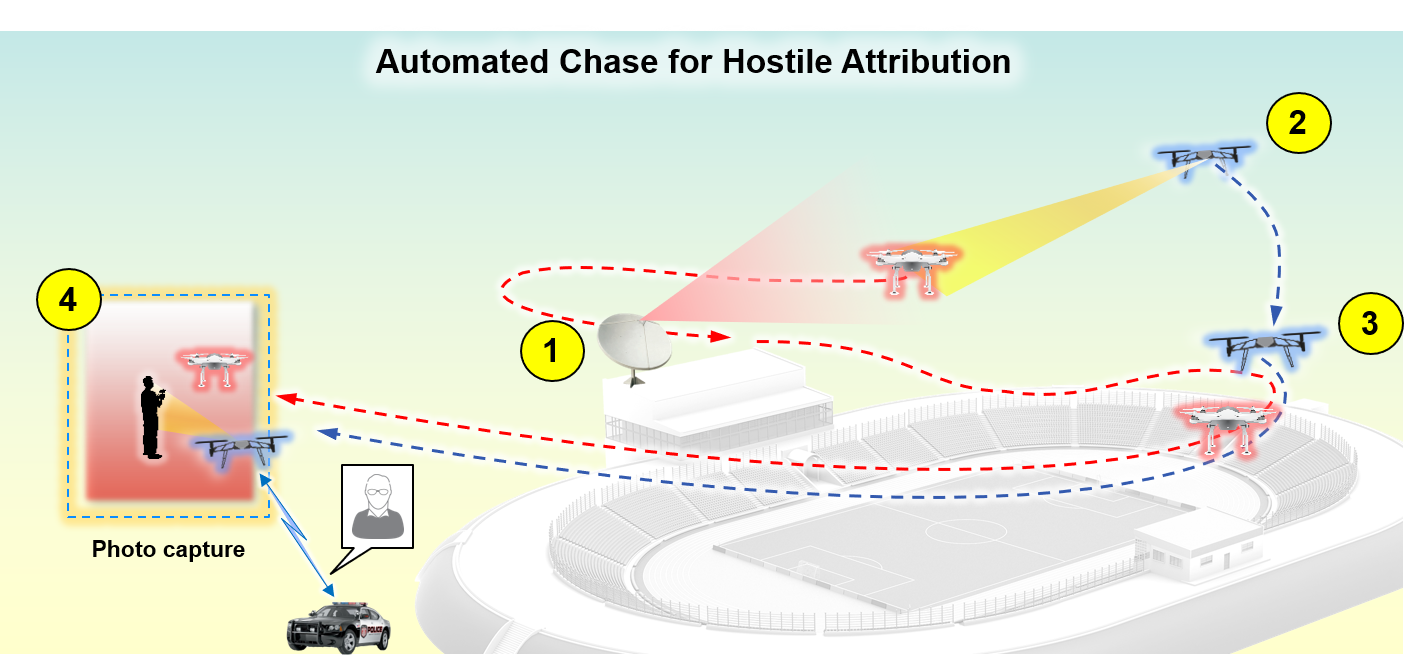}

 		\caption{\textbf{Execution steps for the Autonomous Chase.}}
 		\label{fig:chaseOV1}
 	}
\end{figure}

While we have built the system as a modular architecture to support many missions, in order to demonstrate concretely the potential for an autonomous sUAS, we opted to build out and test in live flights one of the proposed missions: the autonomous chase of an intruding sUAS, as illustrated in Figure \ref{fig:chaseOV1}. Completion of this mission consists of executing the following tasks:

\begin{enumerate}
    \item remote launch on reception of radar cue
    \item flight to cue and target acquisition with on-board sensors
    \item follow target from on-board sensor input
    \item stream information back to agents on the ground
\end{enumerate}

 Each functional block listed above requires the use of multiple hardware or software modules, linked together to execute the full chain of actions and fulfill the mission.

\subsection{Accelerating Development}
With a long-term vision of supporting many different and varied missions, which may share some core autonomy pieces but will need different peripherals to accommodate different specific missions, sensors, etc., we took the approach of developing a modular testbed. We have chosen hardware and platforms that are capable of carrying and powering a wide range of sensors, with robust on-board compute hardware. The flight controls enable several autonomous or semi-autonomous modes, building on Ardupilot flight modes, like Alt Hold and Guided, with additional features like intelligent waypoint generation and navigation, as well as radar cued computer vision search. 

On the software side, we built the detection, tracking, and controls software as individual, modular components, so that each function can be reused for future missions. For example, we expect that modules such as ``follow target from global coordinates'' will be broadly useful to many missions, whereas ``EO computer vision object detection'' may only be useful for a subset of missions. We envision this work forming a foundation of complex sUAS autonomous capabilities that can be adapted and developed to fit future mission-specific needs of agencies and small teams. The modular library of tools developed here helps close the gap between commercial sUAS autonomous features and the needs of small teams for complex sensor-guided autonomous controls. 

The rest of the paper is organized as follows: Section \ref{sec:arch} covers the hardware and software architectures that support the autonomous algorithms, Section \ref{sec:algos} covers the suite of algorithms that enable autonomous flight, including computer vision object detection, motion controls, autonomous search for on-board acquisition including radar/GPS track correlation, and path planning, Section \ref{sec:testing} covers testing results from both simulation and live flights, Section \ref{sec:discussion} addresses various issues and lessons learned from testing, and finally we conclude in Section \ref{sec:conclusion}.

\section{CHASER Architecture}
\label{sec:arch}
\subsection{Hardware Architecture}\label{sec:hardware}
CHASER is capable of carrying a variety of on-board sensors and compute boards. In terms of general hardware required for flight, the sUAS is composed of a DJI S1000 octocopter frame, RFD900x telemetry radio, Raspberry Pi, Pixhawk autopilot, remote control receiver, and Here+ GPS. In addition to the required flight hardware, we use an e-con 4K camera mounted on a 2-axis gimbal, Connex video streamer, and NVIDIA Jetson AGX Xavier board to run all automation and computer vision algorithms. The ground control station uses a Dell XPS laptop and an Echodyne EchoGuard radar.

\begin{figure}
\centering
\includegraphics[width=3.25in]{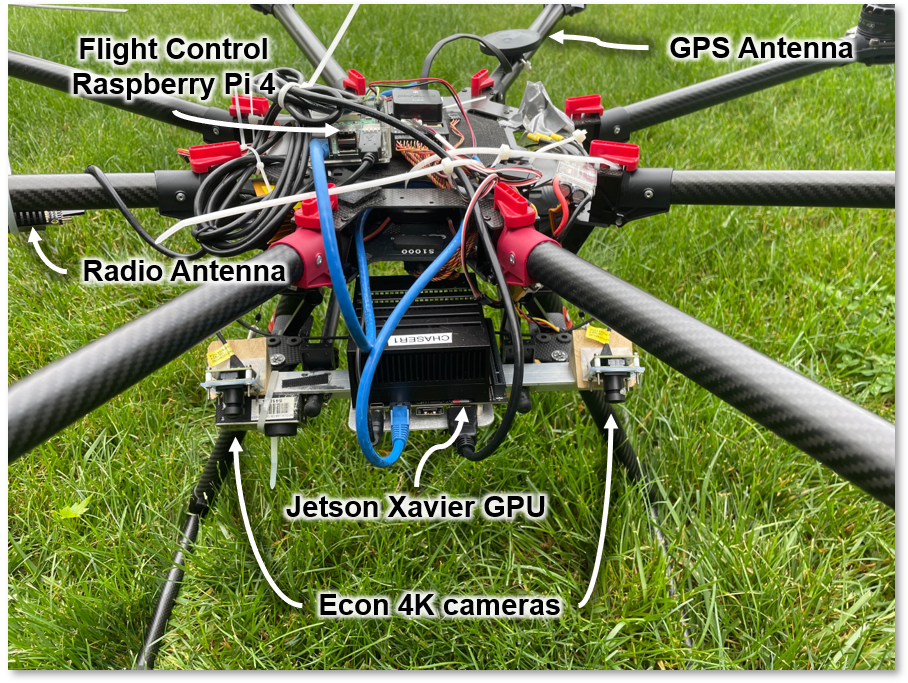}\\
\caption{\textbf{DJI S1000 chase sUAS with custom hardware mounted in preparation of flights.}}
\label{fig:CHASER_yr1_photo}
\end{figure}

The system was originally developed with most of the hardware integrated on board the sUAS, with the exception of the ground control laptop and the EchoGuard, as seen in Figure \ref{fig:CHASER_yr1_photo}. Later, due to changing U.S. regulations, the NVIDIA Jetson AGX Xavier board was debarred from the on-board sensor and processor array. The NVIDIA board was relocated to ground control station while a secondary Raspberry Pi was swapped in its place to host the e-con camera. To provide the NVIDIA board with live video and ground control telemetry streams, a Wi-Fi antenna module and a directional Wi-Fi receiver were attached to the secondary Raspberry Pi and to the ground control local area network, respectively (see Figure \ref{fig:chaserDiagram}).

\squeezeup

\subsection{Software Architecture}
\begin{figure*}
\centering
\includegraphics[width=\textwidth]{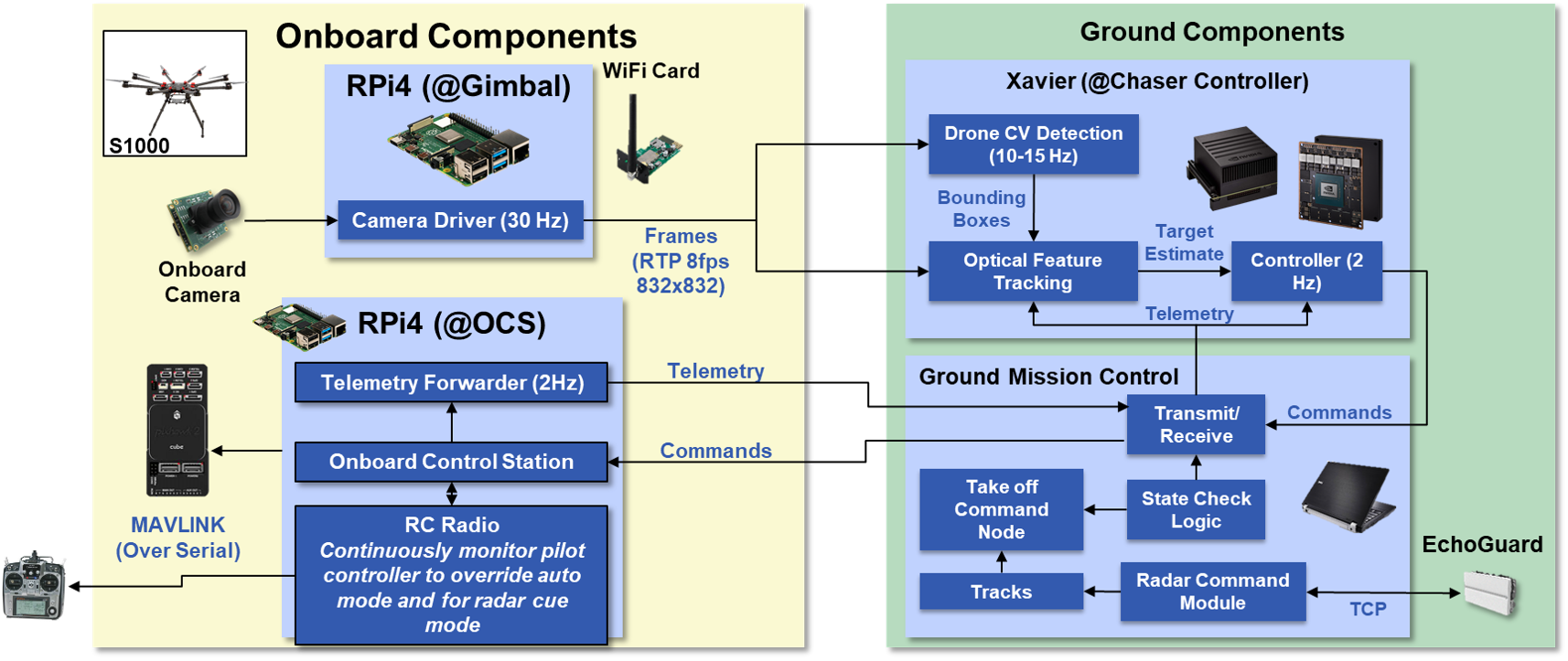}\\
\caption{\textbf{System-level diagram of CHASER, demonstrating the federated system approach with ground system performing computer vision operation and sending commands to airborne command software.}}
\label{fig:chaserDiagram}
\end{figure*}

The architecture of CHASER allows the sUAS to be completely autonomous with all systems on board the aircraft, or function as a federated system with on-board and ground systems working together. Regardless of the architecture scheme, the ground control station can overwrite all commands if desired and the sUAS will return to launch in the event of a loss of communication with the ground station. With a modular architecture in mind, CHASER uses the robot operating system (ROS) to develop control modules, otherwise known as ROS nodes, that can communicate within the local network. As seen in Figure \ref{fig:chaserDiagram}, ROS nodes were developed to run the EchoGuard ground radar, follow controller, detector and trackers, and state machine controller. In this federated configuration, these ROS nodes all communicate with the ground control station through User Datagram Protocol (UDP) message streams to relay message between ground and airborne nodes. On board the aircraft a Raspberry Pi controls the message flow to the Pixhawk autopilot. For all operations CHASER leverages the open source autopilot software ArduPilot, specifically the ArduCopter sub-variant, to maintain flight dynamics of the sUAS through the various mission states. As a safety precaution this Raspberry Pi connected to the Pixhawk also monitors the RC channels set by the pilot in command to determine if the automated function should be followed.

\subsubsection{State Machine}
CHASER uses a state machine that dictates which autonomous behavior is allowed for a given phase of the mission. Overall the system states are:

\begin{enumerate}
    \item On ground -- awaiting radar or system commands
    \item Radar validated track received -- arm and takeoff to predefined target altitude
    \item Radar cue available -- fly out to cue location
    \item Radar cue unavailable -- use last known location to start search
    \item Computer vision acquired -- maintain threat sUAS in field of view
\end{enumerate}

When in state (1) the system requires a validated sUAS radar track to transition to state (2). This validation of the sUAS radar track can be manually confirmed by a human operator or based on the confidence a monitoring system has on the radar track. As the aircraft transitions to state (2) the information from the radar can update the latitude and longitude to fly towards, but will not transition to state (3) or (4) until the target altitude has been achieved. In the event that threat sUAS radar information is no longer available (e.g., sUAS moves out of the field of view of the radar), the system transitions to state (4) and begins a receding horizon search. While in state (3) or (4) the computer vision algorithm is continuously processing camera imagery for detection of threat sUAS and transition to state (5). In the event the threat sUAS moves out of field of view of the on-board cameras CHASER falls back to state (3) or (4).

\begin{figure}[H]
	\centering
	{
		\includegraphics[width=3.25in]{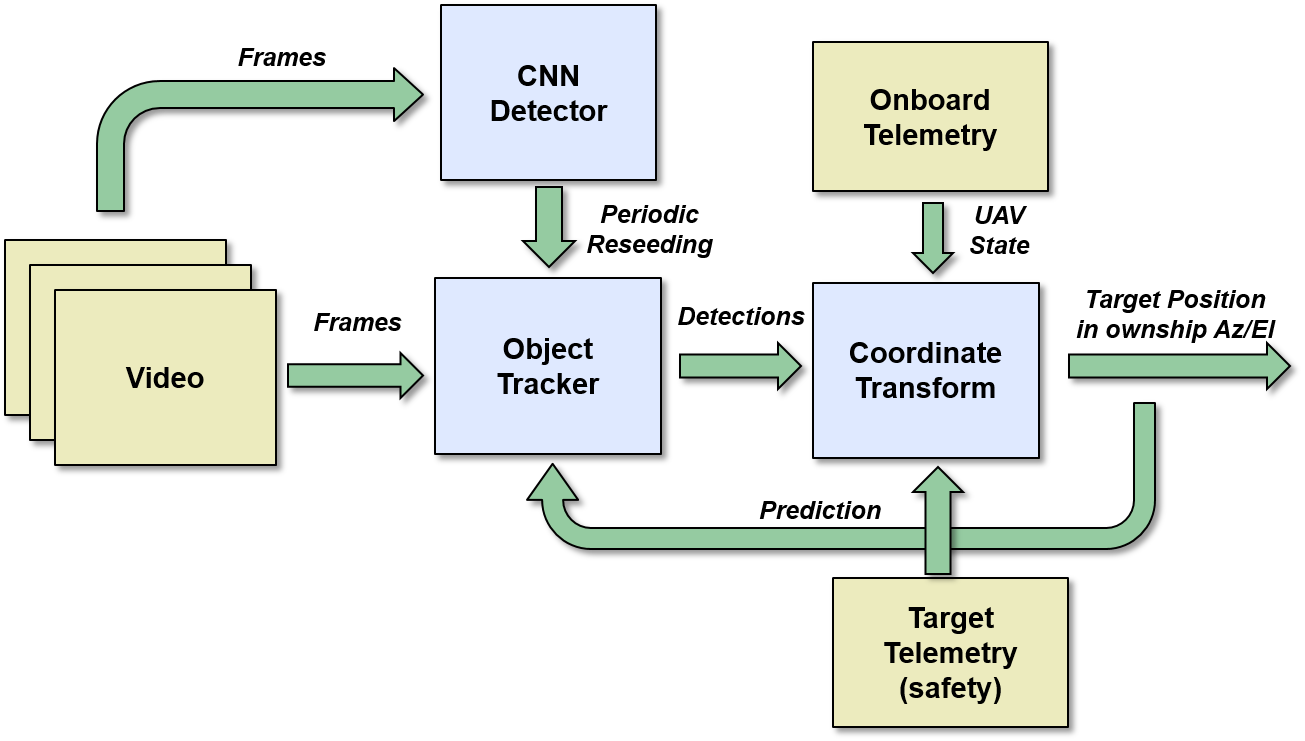}
		\caption{\textbf{Detection and tracking architecture. Data inputs in brown, functional blocks in blue.}}
		\label{fig:tracker}
	}
\end{figure}

\section{Algorithms}
\label{sec:algos}

\subsection{On-Board Detection and Tracking}

We decompose our Detection and Tracking module into three major functional blocks along with three data inputs (see Figure \ref{fig:tracker}). The three functional blocks are:
\begin{enumerate}
    \item Object Detector
    \item Visual Tracker
    \item Coordinate Transformation
\end{enumerate}

The three data inputs are:
\begin{enumerate}
    \item Video frames
    \item Ownship telemetry
    \item Target telemetry
\end{enumerate}

In an operational setting, only the video and ownship telemetry inputs would be available. The target telemetry input is included during development of CHASER both to aid development by separating controls performance from computer vision performance, and for additional safety checks to prevent the Chase sUAS from accidentally flying out of range of the ground control station and crashing.

\subsubsection{Computer Vision Object Detection}\label{subsec:cv}

For acquisition of threat sUAS, we utilize the YOLOv2 architecture for our computer vision object detection model. YOLOv2 is a convolutional neural network (CNN) based general object detector. An object detector is a class of machine learning models that outputs boundary image coordinates (bounding box), classification identification (class ID), and confidence score (from 0 to 1, where 0 is least confident and 1 is most confident) for each predicted object in a given image \cite{yolov2}. General object detectors are utilized to generate predictions in a scene for a wide variety of everyday objects.

In order to adapt the YOLOv2 architecture for our own use case of detecting sUAS, we collected and annotated our own dataset. Data of the S1000 target sUAS was incrementally collected during fields tests and then annotated using the Computer Vision Annotation Tool (CVAT) \cite{cvat2019}. See Appendix \ref{appendix:DataSheet} for more information about our dataset as a datasheet \cite{datasheets}.

We developed our model in the Darknet neural network framework \cite{yolov4}. Before training, we initialized model weights with darknet19\textunderscore448.conv.23 pretrained weights in order to leverage transfer learning from the general object detection setting. During training, we unfroze all layers and performed fine-tuning over the entire network using batches of data sampled from our custom dataset. We selected standard parameters for learning schedule, optimization technique, data augmentation, etc. (see Appendix \ref{appendix:ModelCard} for details).

During live testing, we pulled image frames from the camera feed, performed light pre-processing to form an image tensor, and then queried our model with the tensor to make object detection predictions in real time. Post-processing and thresholds formulated around class ID and confidence scores were used to filter predictions; these filtered predictions were then passed along to the next functional block for further algorithmic processing. For additional details about our trained object detection model, including metrics on its performance, please refer to Appendix \ref{appendix:ModelCard} for our model card \cite{Mitchell_2019}.

\subsubsection{Visual Tracking}\label{subsec:tracking}

In order to improve robustness in computer vision target prediction, we additionally utilize visual trackers. The on-board camera system provides video frames to the CNN object detector, previously described, responsible for detecting a sUAS within the video feed. Because the CNN detector is computationally expensive, it only runs about 8-10 frames per second (FPS). In order to bring target acquisition above 30 FPS, all predicted detections are handed off to a visual object tracker that is lighter-weight computationally and serves to provide continuity between detections from the CNN detector.

The system can use zero or more visual trackers simultaneously. Example options include the Kernelized Correlation Filter (KCF) \cite{kcf_tracker}, the Minimum Output Sum of Squared Error (MOSSE) tracker \cite{mosse_tracker}, the Generic Object Tracking Using Regression Networks (GOTURN) tracker \cite{goturn_tracker}, and the Tracking Learning and Detection (TLD) tracker \cite{tld_tracker}. Since these visual trackers are incapable of performing object detection themselves, they need to be reseeded periodically with detections from the CNN detector algorithm. Detections from the CNN object detector are also utilized to ensure existing tracks are reasonable. For example, a visual tracker can drift off of the original detected object and instead follow something such as a bush or shrub on the ground. This can be avoided by pruning tracks maintained by the visual tracker that do not align with any detection from the CNN object detector.

\subsubsection{Coordinate Transformation}\label{subsec:coord_transform}
Finally, target estimates (namely, the bounding boxes) from the detector or tracker are maintained in pixel space and require a transformation into real-world 3D coordinates before processing by the controls algorithm. The coordinate transformation from pixel space to real-world 3D space requires knowledge of 1) the ownship orientation relative to a fixed east-north-up (ENU) plane, 2) the translation and orientation of the camera relative to the center of the ownship, 3) the camera field-of-view (FOV), and 4) the resolution of the video feed. While 2-4 are static and known apriori, ownship orientation is dynamic and is provided by an on-board Intertial Measurement Unit (IMU). Ownship orientation is processed through an alpha-beta tracker to reduce measurement noise and smooth out rapid accelerations prior to being used in the coordinate transformation algorithm. Due to the wide field of view of the camera and high mounted elevation, we implemented transforms between coordinate reference frames using quaternion rotations.

\subsection{Relative Position Controls} \label{sec:follow}

\begin{figure*}
\centering
\includegraphics[width=\textwidth]{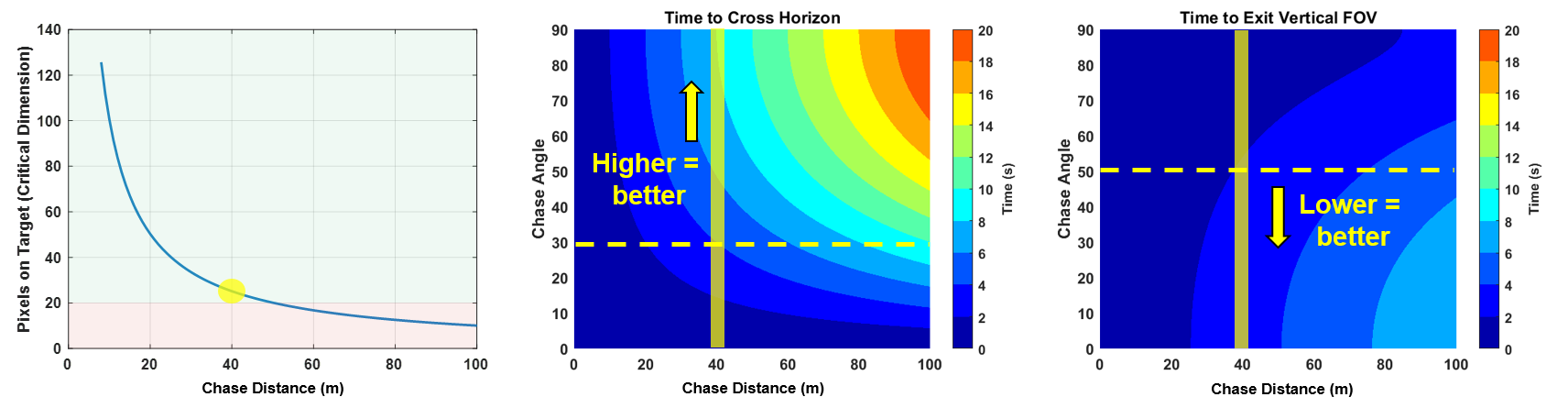}
\caption{\bf{Analysis of optimal chase position. Left: determining sufficient pixels on target to support CNN-based detection. Center: Given required number of pixels on target, selecting a chase angle that optimizes for a consistent background. Right: Given the required number of pixels on target, selecting a chase angle that gives maximum resilience to sudden target motion.}}
\label{fig:chaseanalysis}
\end{figure*}

To autonomously follow a target position, such as one computed by our detection and tracking module, we developed relative position controls which allow for following at a specified offset. From the input, relative position coordinates with respect to CHASER, the algorithm computes the required yaw rate, climb rate, and acceleration for the platform to maintain a set position. We designed our controls algorithm to work with a fixed camera (non-gimbaled), and we performed systems analysis of the sUAS-chase mission to choose a set positional offset that would give the highest chances of maintaining the target in the FOV. The results from that analysis are detailed here and in Figure \ref{fig:chaseanalysis}.


The controls algorithm is designed to work with ArduCopter's Alt Hold flight mode. In this mode, the ArduCopter autopilot system maintains the current altitude when no vertical climb rate commands are given. The commands that can be  received in this mode are roll, pitch, and yaw rates, along with the vertical climb rate. In Alt Hold mode, ArduCopter's autopilot software handles all of the changes in rotor speed required to achieve a desired roll, pitch, yaw, and climb rate, so we can just compute the desired rate inputs using our algorithm. 

We elected to design our controls to command 3 of the 4 possible ArduCopter Alt Hold inputs: vertical climb, pitch, and yaw rates. Roll was considered a redundant command for our initial implementation; the one axis of horizontal motion allowed by pitch commands (forward/reverse) could be rotated $\pm90^{\circ}$ with heading changes from yaw commands to cover the entire 2D plane of horizontal motion. However, we recognize that roll commands are needed to take full advantage of the agility of multicopters, and so we leave their implementation in the algorithm to future work.

At every step of the controls loop, we have two objectives: 

\begin{enumerate}
    \item Maintain the target in the FOV of the camera for continued detection and tracking.
    \item Limit target range to allow for continued detection.
\end{enumerate}

Note, these two objectives are not independent with a fixed camera, that is they share control variables. For instance, to accomplish objective (1), the platform needed to be pitched and yawed in the direction of the target. However, to accomplish objective (2), the platform needed to pitch or climb to reach the desired range. The common control variable means that at times, these two objectives would interfere with one another. For instance, if the target suddenly moves downrange of CHASER (flying above the target), the target moves upward in the FOV, requiring that CHASER pitch backward to accomplish objective (1). Although, to accomplish objective (2) CHASER must pitch forward. Therefore, during controller tuning it must be considered that these objectives at times may be at odds with one another.

To determine the best position from which to follow the target and accomplish these objectives, an analysis was performed of chase position considering the dynamics of a DJI Phantom 4 quadcopter (a popular consumer quadcopter) as the target vehicle. The analysis first looked to constrain the volume of following positions by considering the benefits of flying autonomously above or below the target. Given advantages in safety of flight path and background consistency, we elected to fly above the target. We also chose to follow the target from behind, to avoid detection in the target's camera, and to better respond to changes in motion. To determine the optimal elevation angle and slant range for our chase vehicle with respect to the target, we asked three guiding questions for a given chase position: 
\begin{enumerate}
    \item How many pixels are on target (critical dimension pixels, $\sqrt{\texttt{bbox}_w*\texttt{bbox}_h}$)?
    \item How much time to react to vertical maneuver before target crosses horizon line assuming target instantaneously moves at a maximum vertical speed?
    \item Assuming target instantaneously moves at a maximum horizontal speed, how much time to react before target leaves the vertical FOV?
\end{enumerate}


From this analysis, we determined that the position to follow a target from would be 40m slant range, 40$^{\circ}$ elevation (above the target), and with the fixed camera tilted downward at the angle of elevation above the target. However, we noted that ranges between 30-50m and elevations between 30-50$^{\circ}$ would likely be sufficient for a sustained chase, as seen in Figure \ref{fig:chaseanalysis}. This choice in slant range therefore becomes the objective of our controls loop for the chase vehicle. Given the pitch and climb rate inputs to ArduCopter's Alt Hold mode, it made sense to convert our slant range and elevation angle to optimal relative height and ground range setpoints to simplify the control loop formulation.

Our controls loop was designed with 3 Proportional-Integral-Derivative (PID) loops to control the three command variables independently -- yaw rate, pitch, and clime rate. The PID coefficients are described in Table \ref{tab:pid_coeff}.
\begin{table}
\renewcommand{\arraystretch}{1.3}
\caption{\bf Relative Position Control PID Coefficients}
\label{tab:pid_coeff}
\centering
\begin{tabular}{|c|c|c|c|c|c|}
\hline
\bfseries Control  & \bfseries Process  & \bfseries Set Pt. & \bfseries $K_P$& \bfseries$K_I$ & \bfseries$K_D$\\
\hline\hline
Yaw Rate & Az. & 0$^{\circ}$ & 0.25 & 0.002 & var \\
Pitch & Gr Rng & 30.6m & 0.4 & 0.01 & 0.2 \\ 
Climb Rate & Rel. Ht. & 25.7m & 0.6 & 0.01 & 0.02 \\
\hline
\end{tabular}
\end{table}


The controls loop was tuned to run at 5 Hz through simulation in the SITL environment. The Yaw Rate PID controller has a $K_D$ value that varies linearly with target ground range, as $K_D = 0.0003R$, to account for more rapid target azimuth changes at shorter ranges. Given noise in inputs from the tracker, a median filter was added to the controls algorithm, filtering the process variables with a window of 5 measurements. These three PID loops work together to position CHASER at the desired setpoint with respect to the target. The algorithm was implemented in Python and a number of mechanisms were employed to ensure the safety of the CHASER system. On exit, the control algorithm always sends a hover command to the platform, zeroing out any previous commands and causing the system to stop motion. The same hover command is also sent after the positions from the tracker become ``stale," or the tracker fails to send positions to the controls for a set amount of time (2 seconds).

\subsection{Radar Track Correlator}
\label{sec:trackcorr}
For more effective use of the ground-based EchoGuard radar, we developed a radar track correlation algorithm, capable of sifting through the tracks of multiple objects in the radar FOV, and associating tracks with known vehicles. Without such correlation, we cannot differentiate our ``ownship" track from the radar from any foreign sUAS radar tracks, rendering its output ambiguous as soon as CHASER enters the radar's FOV. Furthermore, although the radar is capable of assigning a ``Track ID" to a series of continuous detections on a moving target, this ID may not be consistent across the entire duration of the target's flight.

Using the CHASER platform's on-board GPS, we can develop an algorithm to associate a track from the radar with our ``ownship" position, and filter that track from the radar output, leaving only unassociated tracks for additional processing. We designed our track correlation algorithm to store the Track ID from the radar track associated with the GPS position, so that we could filter on Track ID alone. To do so, we needed to account for known failure modes of the radar Track ID system, including:

\begin{itemize}
    \item Track dropping: an object's track suddenly ends, but the radar picks it up with a new Track ID
    \item Track swapping: when two objects with radar tracks approach each other, their tracks may ``swap", leading to the associated Track ID changing.
    \item Duplicate tracks: the radar may generate parallel tracks for a singular object
\end{itemize}

A way to recover from track dropping and swapping is to continuously verify if the stored Associated ID is still valid for the known, GPS-tracked, object. To avoid multiple tracks meeting a validity criterion, there must be a definite condition to select a singular radar track ID. The following algorithm takes all of these into account.

The GPS data from the known ``ownship" position and incoming radar track data are rotated to a common coordinate system and compared to see which of the tracks consistently falls close to the GPS position, which would identify the associated track. For each active track, the GPS data is temporally interpolated to match the active track's timestamps, and individual timestamps are compared for valid detections, where the GPS and radar track positions fall within some distance threshold, $\epsilon$. The number of valid detections are counted for different windows of time. First, the most recent two second window is checked for track similarity to GPS positions, and the best track's ID may be selected for as the new associated track ID. Under certain conditions, the algorithm may perform an additional evaluation, examining older detections falling into timestamp windows ($T-4,T-2$] and ($T-6,T-4$], indicating seconds from current time $T$. Under these conditions, if a track achieves the most detections in more than one of these windows, its ID is chosen the winner. See details in Algorithm \ref{alg:correlation}. 

\begin{algorithm}
\caption{Radar Track Correlation}\label{alg:correlation}
\begin{algorithmic}
\State $\{R_1, R_2, ...\}$ \Comment{Set of Active Radar Tracks}
\State $G$ \Comment{GPS track}
\State $T$ \Comment{Current Time}
\State $\epsilon$ \Comment{Detection Threshold Distance}
\State $A \gets \texttt{null}$ \Comment{Associated Track ID}
\For{$i$ in $\{1,2,..\}$}
    \State $\tilde{G_i} \gets $ \texttt{interpolate($G$, $R_i$)}
    \State $N_i^{(0)} \gets \mathlarger{\sum}_{t>T-2}(|R_i^{(t)} - \tilde{G_i^{(t)}}| < \epsilon)$ \Comment{Detections}
\EndFor

\If{$A$ is \texttt{null}}
    \State $A \gets \mathlarger{\argmax}_{i}$  $N_i^{(0)}$ \Comment{Track w/most detections}
\Else
\If{not $A == \mathlarger{\argmax}_{i}$  $N_i^{(0)}$}
    \For{$i$ in $\{1,2,..\}$}
        \State $N_i^{(1)} \gets \mathlarger{\sum}_{T-2\geq t>T-4}(|R_i^{(t)} - \tilde{G_i^{(t)}}| < \epsilon)$ 
        \State $N_i^{(2)} \gets \mathlarger{\sum}_{T-4\geq t>T-6}(|R_i^{(t)} - \tilde{G_i^{(t)}}| < \epsilon)$ 
    \EndFor
    \State $m^{(j)} \gets \mathlarger{\argmax}_{i}$  $N_i^{(j)}$ for j in \{0, 1, 2\}
    \If{any of $m^{(\{0,1,2\})}$ equal} 
    \State $A \gets $ most frequent element of $m$
    \EndIf
\EndIf
\EndIf

\end{algorithmic}
\end{algorithm}

Addressing the expected failure modes of the radar, the algorithm continuously evaluates for the most detections over time windows, so it can catch track ID dropping and swapping errors throughout the flight and recover an object's current associated ID. Additionally, by choosing the radar track with the most detections within the distance threshold, the algorithm has a consistent way to pick from duplicate tracks.

\subsection{Receding Horizon Controller}
\label{sec:rhc}

\begin{figure*}[t!]
\centering
    \subfigure[  Before]{
        \includegraphics[width=0.27\textwidth]{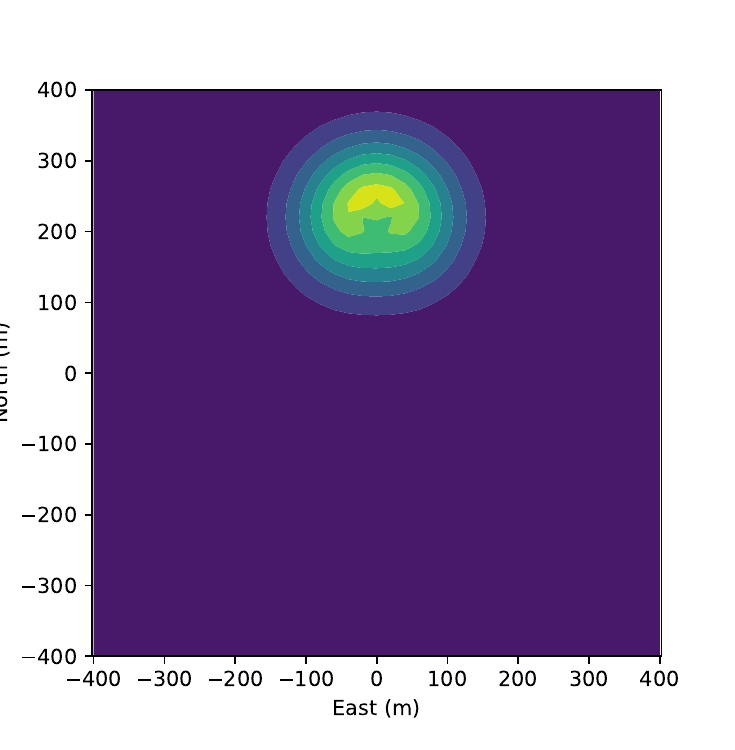}
        \label{fig:RHCtripletA}}
    \subfigure[  Path Generation]{
        \includegraphics[width=0.36\textwidth]{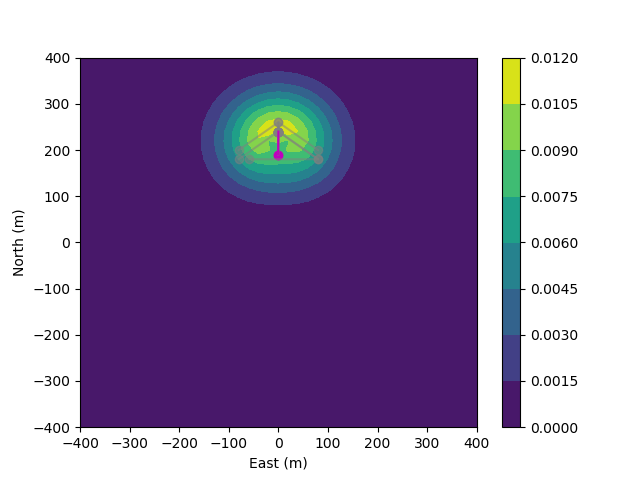}
        \label{fig:RHCtripletB}}
    \subfigure[  After]{
        \includegraphics[width=0.33\textwidth]{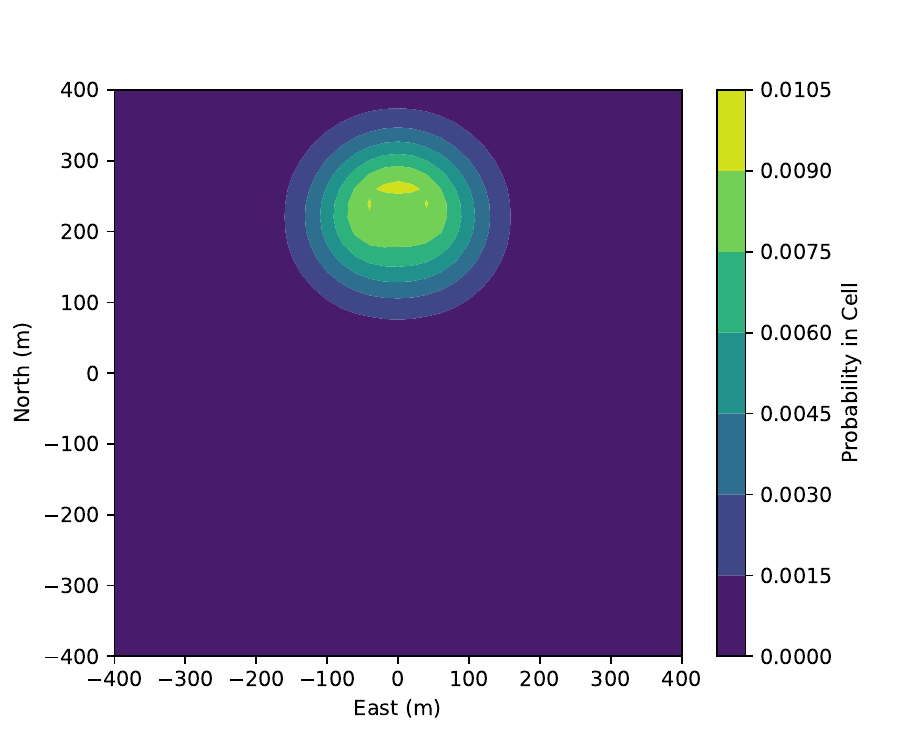}
        \label{fig:RHCtripletC}}

    \caption{\textbf{Visualization of the RHC algorithm, with updates to the belief map based on CHASER motion. In A, a belief map is created based on past belief, target dynamics, and the sensor model for the CHASER platform. In B, five potential paths are generated for CHASER, with the first step of the best path highlighted in magenta and subsequent steps for each path in gray. In C, after Chaser follows the command to move in the direction specified in B, the belief map is updated by target dynamics and CHASER's sensor model}}
    \label{fig:data_aug}
    \vspace{-0.5cm}
\end{figure*}

Through use of the Radar Track Correlator, we can continuously receive information on the radar tracks that are not associated with CHASER, allowing for cueing to the target's most recently detected position. However, if the radar should lose its track on target, large radar detection uncertainties were to develop, or the radar itself were to fail, our vehicle would need a way to effectively search the volume of airspace around the singular target's most recent position with the goal of establishing vision-guided follow control. To approach this problem, we implement a Receding Horizon Controller (RHC) \cite{kwon_han_2006} to optimize path planning over the possible positions of the target (referred to as the target belief map).

We approached the setup to the RHC by creating a target dynamics model, a sensing model, and a belief model, similar to the approach of \textit{Leahy \& Schwager, 2018} \cite{leahy2018}. In this arrangement, the target dynamics model defines the possible state transitions of the target vehicle, the sensing model defines the expected observability of the environment using CHASER's camera, and the belief model, initialized to the last known position of the target, keeps track of likely target positions through interactions with the target dynamics and sensing models. 

To discretize the setup of RHC and our belief model, we create a uniform three-dimensional grid of points $Q$, where each point on the grid $q$ represents a state that can be occupied by the target or by CHASER. We estimate the probability of the target occupying the grid cell surrounding a point $q$ as $b$, with $B$ representing the collection of all such $b$ that make up the full belief model. Both the target dynamics and sensor models operate on $Q$ to affect the belief model $B$. At time $t_0$, we have an initial belief about the position of the target $B_0$, based on the radar's last detection and that detection's uncertainty in range, azimuth, and elevation. At each update to the belief map after elapsed time $\textbf{dt}$, we approximate a target's state transitions by a 3D Gaussian probability density function (PDF), with $\sigma_x = \sigma_y = v_{h}\textbf{dt}$ and $\sigma_z = v_{v}\textbf{dt}$, where $v_h$ and $v_v$ are target's maximum expected horizontal and vertical speeds. This becomes our target dynamics model. As such, the target has some probability of transition to adjacent states based on its expected speed, elapsed time, and distance to the adjacent state. We compute this update to the belief model $B_t \rightarrow B_{t+1}$ with a 3D convolution over $B_t$, where values of the convolutional kernel are given by the 3D Gaussian transition PDF and distances between the grid points. The effect is that over time, the expected position of target expands from the initial belief maximum, and the estimate of target position becomes smoothed.

At the same time, we use a sensor model $S$ to compute the set of grid points $q_s \in Q$ over which CHASER's computer vision detection algorithm is effective (within field-of-view of camera and within expected range of detection), given the approximate position of CHASER $x_t$ at time $t$ and its yaw. This sensor model is used both during the belief model update phase as $S(x_t, \texttt{yaw})$ to incorporate information gained from our search into the belief map (see Figure \ref{Fig:BeliefMap}), and during the RHC prediction phase as $S(x_t, x_{t-1})$, where the path from $x_{t-1}$ to $x_t$ determines yaw, and the sensor model computes an expected change to the belief model at future times. The sensor model's effect on the belief model is to \textit{reduce} the likelihood of target position at points $q_s \in Q$ to account for CHASER's recent observation of those points.

In the RHC prediction phase, we aim to choose the path which offers the highest likelihood of acquiring the target using computer vision, based on our current belief map $B_t$. To do so, we use an iterative algorithm (\ref{alg:cap}) to examine all possible paths within some number of steps $h$, and apply the sensor model to compute the total "belief gain" over that path, or sum of belief points $g=\sum\{b_{t+1}^{(1)}:b_{t+h}^{(n)}\}$ computed by the sensor model along the path $ p=\{x_t, x_{t+1}, ... x_{t+h}\}$.

\begin{algorithm}
\caption{Optimal Path on Target Belief Map}\label{alg:cap}
\begin{algorithmic}
\State $B_0$ \Comment{Initial Belief}
\State $h$ \Comment{Horizon}
\State $\gamma$ \Comment{Discount Factor}
\State $p_0 \gets \{x_0\}$ \Comment{Path initialization}
\State $P \gets \{(p_0, g_0)\}$ \Comment{Path, Belief Gain tuple set}
\State $t \gets 1$
\While{$t < h$}
    \For{$p_i, g_i \in P$}
        \State $B_i \gets F(B_0, p_i)$ \Comment Update belief from path
        \State $x_{nbs} \gets \texttt{Adjacent}(p_i[t])$ \Comment{Get neighbors}
        \For{$x \in x_{nbs}$}
            \LineComment{Compute Belief Gain from Sensor Model $S$}
            \State $g_{new} \gets g_i + \gamma^{t}\sum_{q_k \in S(x_t, x_{t-1})}b_{ik}$
            \If{$g_{new} > 0.001$} \Comment{Threshold Gain}
                \State $p_{new} \gets p_i \cup \{x\}$ 
                \State $P \gets P \cup \{(p_{new}, g_{new}\})$ \Comment{Add path}
            \EndIf
        \EndFor
    \EndFor
    \State $t \gets t+1$
\EndWhile
\State \Return{Path, Gain pair with maximum gain in $P$}
\end{algorithmic}
\end{algorithm}

The path that has the maximum likelihood of observing target is returned, and we take just one step along that path before re-running the algorithm to determine the new best-path. If no paths meet the required gain threshold, we simply move to the maximum point on the normalized target probability map. 

\section{Testing}
\label{sec:testing}
To reduce the need for flight testing, the architecture and algorithms were initially tested with a Software-in-the-loop (SITL) using ArduPilot \cite{ardupilotRef}. We were able to test all of our software on a Linux virtual machine (VM) using this simulation tool, which features a physics simulator to emulate flight dynamics of one or more vehicles and the identical autopilot module that would be used during physical tests. After thoroughly testing the software in the simulation environment, we took the physical system described in section \ref{sec:hardware} to a field site where we could test algorithms in real time.

To demonstrate the full system in live flight, we conducted field tests of the CHASER testbed on July 15-16, 2021 at Gardiner Municipal Airport, and July 18, 2022, and August 15-16, 2022 at Fort Devens in Devens, MA. By running our custom software within a ROS framework, we were able to capture all of the data generated during each of the SITL and field tests with synchronized time stamps, and conduct analyses post-flight.

\begin{figure}[ht]
	\centering
	{
		\includegraphics[width=3.25in]{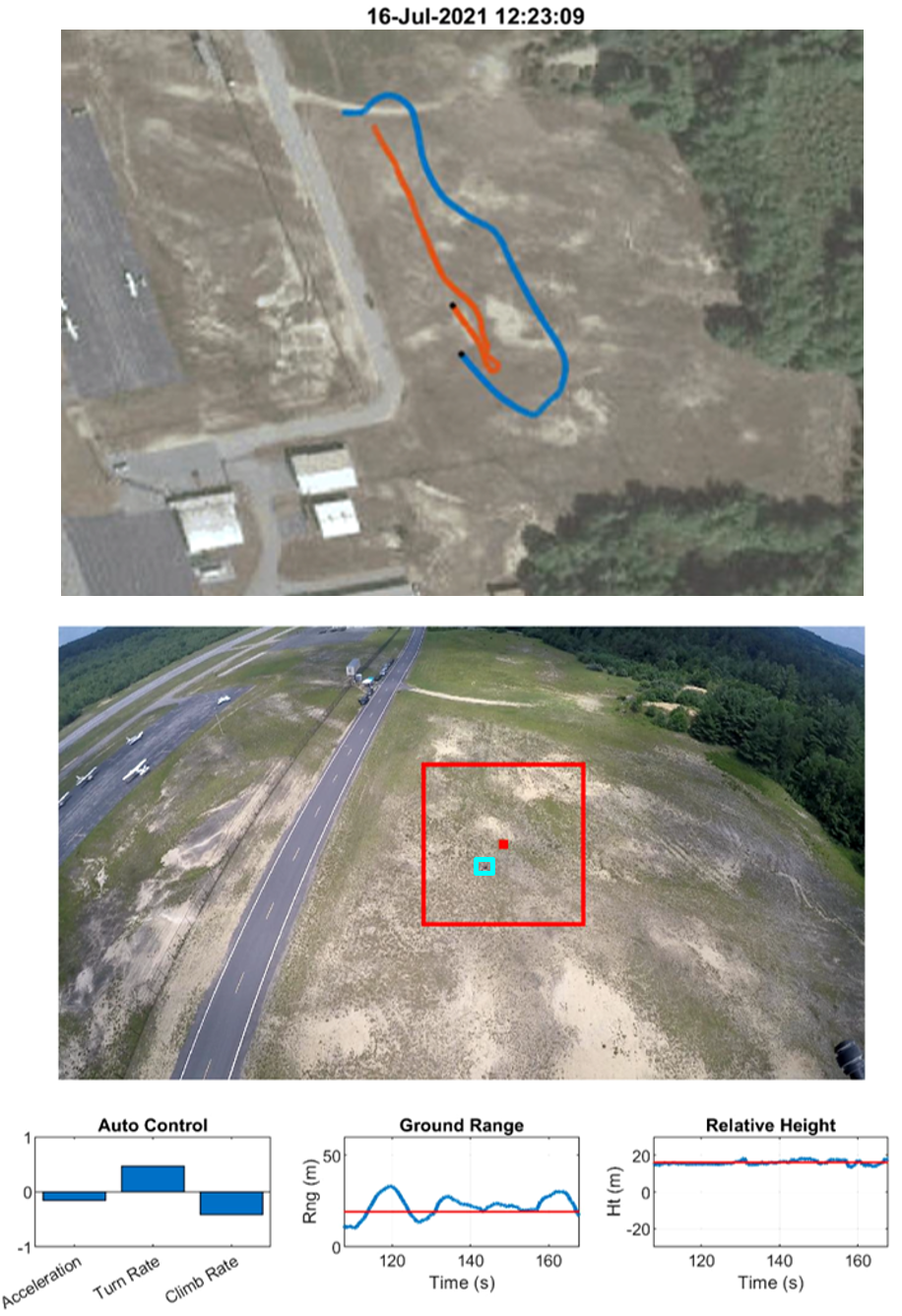}
		\caption{\textbf{A snapshot of the video generated from the July 2021 flight test. Top: overhead view of Target (orange) and CHASER (blue) sUAS flight paths. Middle: Chase sUAS camera view with expected location of target sUAS from telemetry (red) and true position of target (cyan). Bottom Left: control commands sent to the flight controller, Bottom Center: the desired (red), and true (blue) ground range of target rel. CHASER, and Bottom Right: desired and true height of target rel. CHASER }}
		\label{fig:ResultsOverhead}
	}
\end{figure}

\subsection{Computer Vision Detection and Tracking}
In July, 2021, we demonstrated the computer vision detection and tracking algorithms during our live flight test where we also demonstrated the PID controls.

See Figure \ref{fig:ResultsOverhead} for a sample of the on-board camera view with bounding box detection overlaid, and an overhead view of the target and chase sUAS tracks. Post-flight we labeled every frame in the mission video using CVAT (see Section \ref{sec:algos}), and then calculated a precision of $75\%$ and a recall of $44\%$.

\begin{figure}[h]
	\centering
	{
		\includegraphics[width=3.25in]{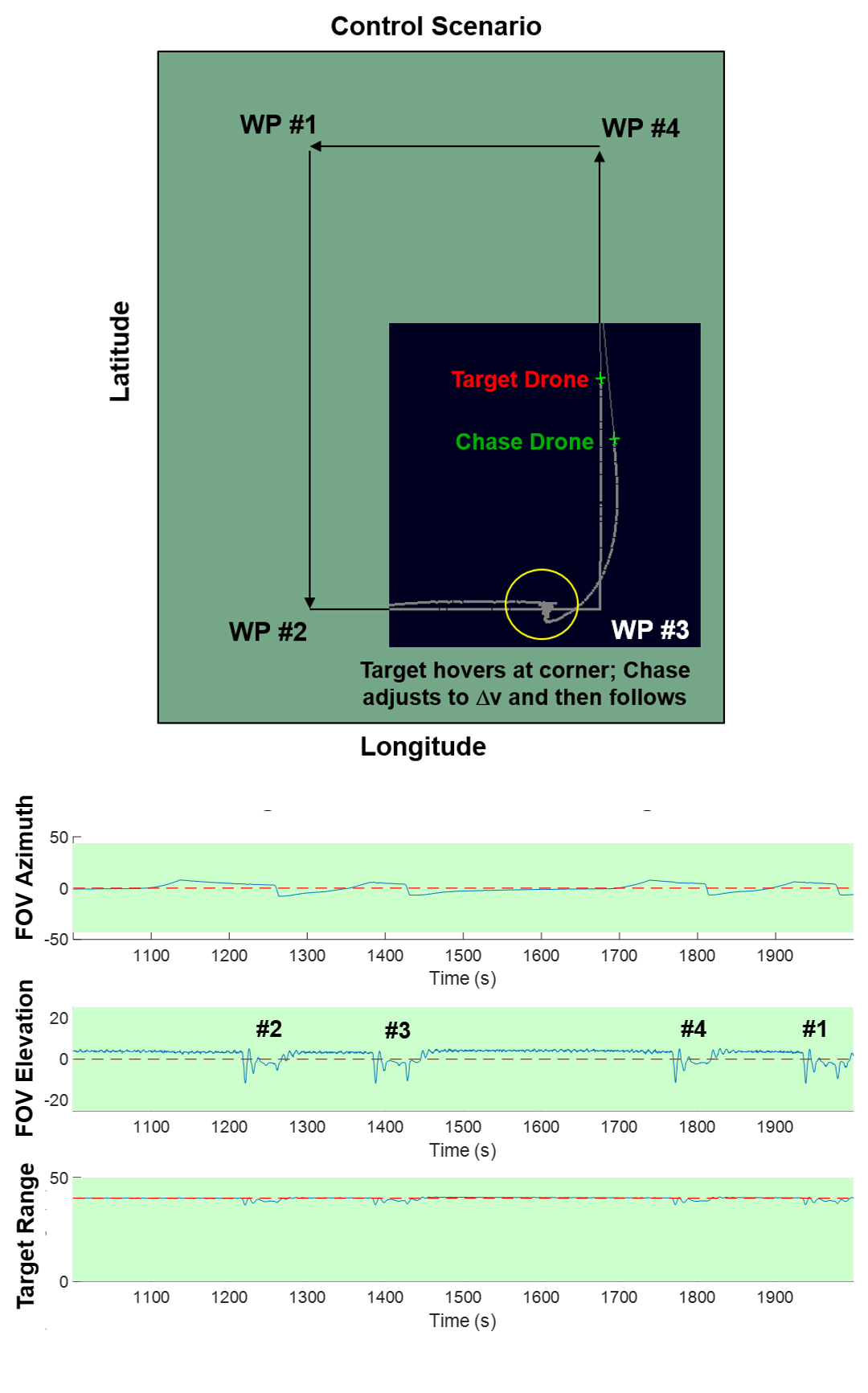}
		\caption{\textbf{Rectangular flight path PID test in SITL, showing the chase sUAS adjusting controls correctly at each corner turn. Top: overhead view of planned flight path, simulation results overlaid in black, showing Chase sUAS adjusting and hovering at corner, then resuming and making the turn to follow Target sUAS. Bottom: Azimuth, Elevation, and Range position of target sUAS relative to chase sUAS over time. Corner turn responses can be seen at t=1250, 1400, 1800, and 1950.  }}
		\label{fig:sitldev}
	}
\end{figure}

\subsection{Relative Position Controls}

We used the physics simulation environment in the SITL to develop and refine the relative position controls algorithm prior to live flight tests. The SITL environment allowed us to simulate a variety of chase scenarios, investigate whether the controls algorithms created the desired following behavior, and tune the PID coefficients under controlled conditions. For example, we set up a rectangular flight path for the target sUAS where it paused at each corner point, and  were able to observe the controls algorithms directing the chase vehicle to appropriately halt, hold its position, and then resume flight and make the turn at each corner point following the target sUAS (see Figure \ref{fig:sitldev}).

\begin{figure}[h]
	\centering
	{
		\includegraphics[width=3.25in]{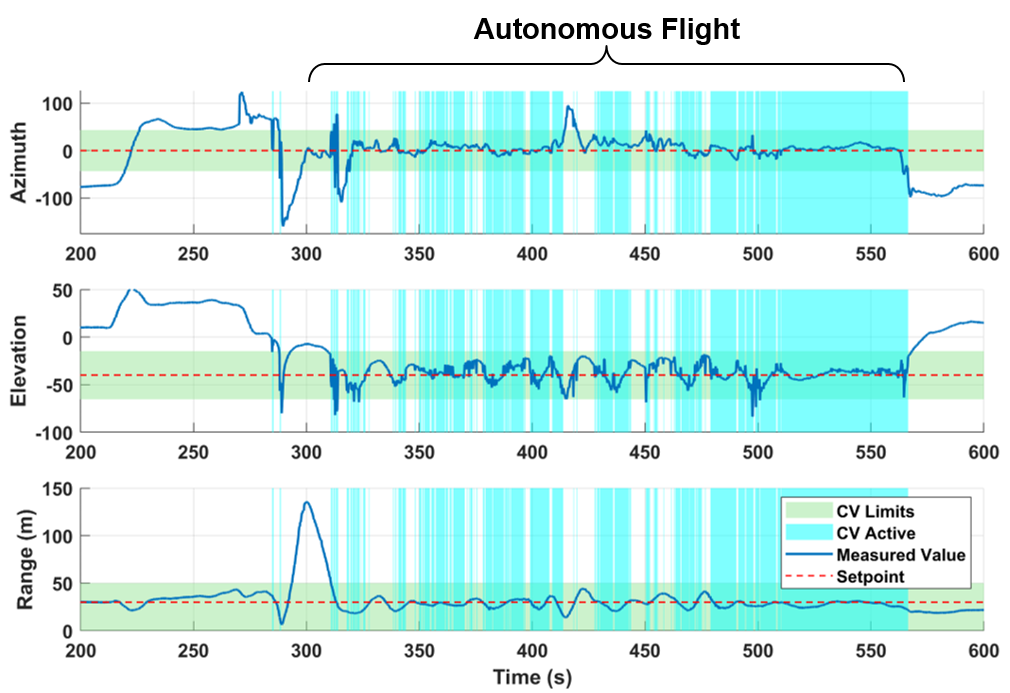}
		\caption{\textbf{Relative Azimuth (deg), Elevation (deg), and Range (m) of Target sUAS to Chase sUAS during July 2021 flight test. Comparison of desired relative position (red dashed line) to true relative position (blue solid line). Green bar indicates FOV and range limits of camera detection, cyan bars indicate intervals when computer vision algorithm is providing detections to controller.}}
		\label{fig:chaseposition}
	}
\end{figure}

In our July 2021 flight test we demonstrated the PID-based relative position controller successfully maintaining the desired position above and to the rear of the target sUAS, operating on a combination of azimuth and elevation data provided by on-board computer vision detections and range provided by GPS (see Figures \ref{fig:ResultsOverhead}, \ref{fig:chaseposition}). The controller maintained CHASER at a position with the target in the field of view for $91\%$ of the total flight time. The computer vision algorithm provided detections on $66\%$ of frames when the CHASER vehicle was in autonomous flight mode (i.e., not including manual piloted take off and landing), as indicated by the cyan shading. 

\subsection{Radar Track Correlator}

To test the radar track correlation algorithm, we flew the CHASER platform along with an additional target sUAS in the field of the view of the radar on Aug 15, 2022. The flight lasted approximately 500 seconds, with both sUASs running circular flight patterns, as well as flying 100m downrange and back. We recorded the output tracks from the calibrated radar and the GPS of CHASER, running our correlation algorithm to filter tracks that were associated with the ownship platform. 

\begin{figure}[h]
	\centering
	{
		\includegraphics[width=3.25in]{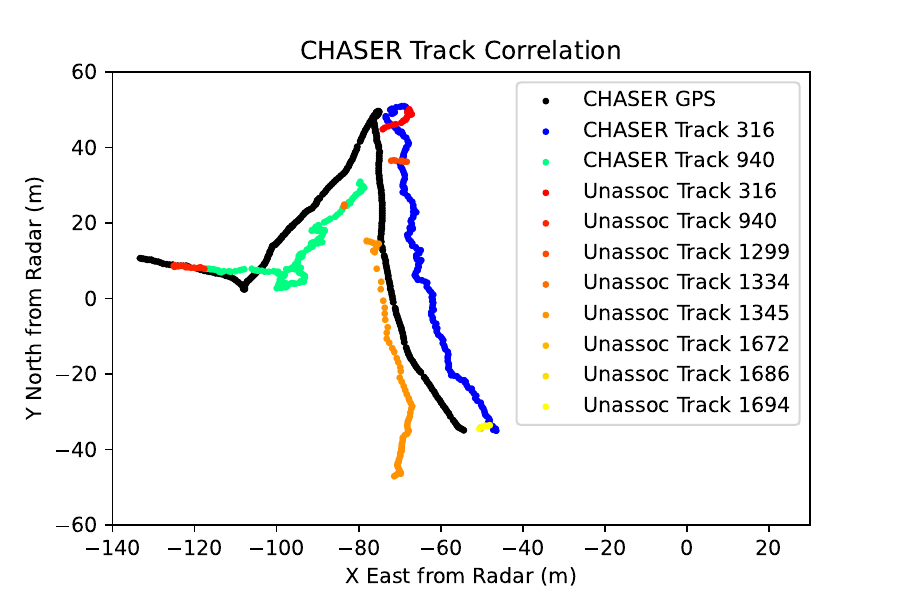}
		\caption{\textbf{GPS position of CHASER (black) between 100-200 seconds after takeoff, with corresponding associated radar track segments (IDs 316 and 940) and unassociated track segments. The direction of the CHASER platform on this flight path was from west to east.}}
		\label{fig:correlation}
	}
\end{figure}

During the flight we observed the 100 second window illustrated in Figure \ref{fig:correlation}, which we found to be representative of the performance of the correlator during this flight. We observe that that the track ID associated with CHASER's position switched from 940 (teal) to 316 (blue) in the middle of this portion of the flight, as CHASER made a sharp turn south. This likely indicates that ID 940 was dropped and a new track was formed. We also observe that associated tracks, upon formation, are typically unassociated for a short time before they are correlated with the GPS position by the algorithm. 

We found that the radar was producing tracks during 99.9\% of the 499.3 second flight, defined by the number of 1 second windows during the flight containing radar tracks. Similarly, we found that during 91.7\% of the flight, the radar track correlation algorithm was associating radar tracks with CHASER.

\subsection{Receding Horizon Controller}
We first ran tests of our receding horizon controller (RHC) in the SITL environment, and compared the RHC performance against a simpler ``maximum belief'' algorithm. For the maximum belief algorithm, we used the same underlying target dynamics model and dynamically updated belief model as was done with the RHC, but we simply cued CHASER directly to the maximum probability point on the belief model, rather than predicting the optimal path over N steps as in RHC. We also varied the discount factor and the horizon of the receding horizon controller to see how these parameters might affect performance. Note that the difference between RHC with a horizon of 1 and the maximum belief algorithm is that the RHC may only take a "step" up to a maximum distance, whereas maximum belief may cue to any point on the belief map.

\begin{figure}
\centering
\includegraphics[width=3.25in]{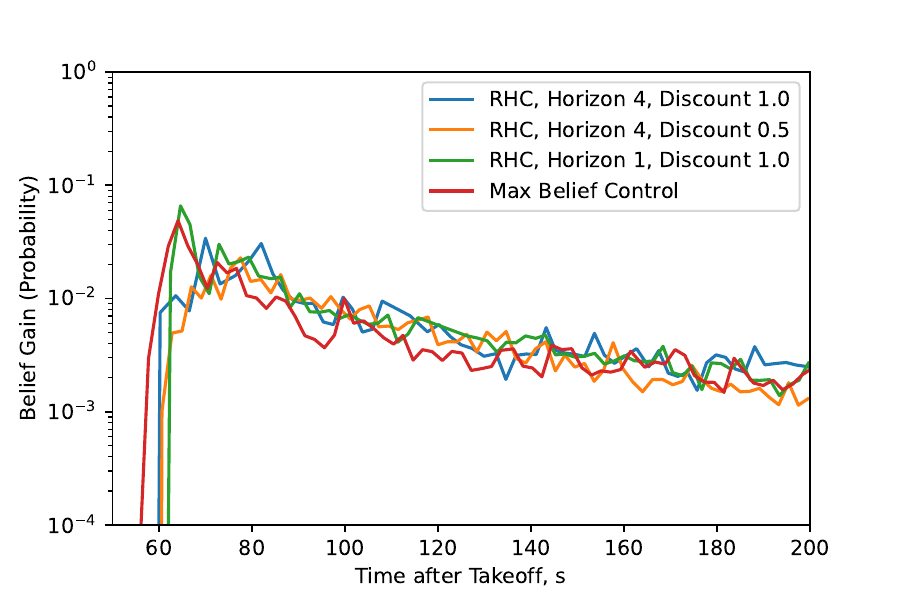}\\
\caption{\textbf{Belief gain in the SITL over time. We compare the Receding Horizon Controller RHC to a baseline algorithm, cueing the CHASER platform to the maximum probability point on the belief map.}}
\label{fig:RHC_sitl}
\end{figure}

Under a simplistic case with no barriers or holes in the potential positions of the target, we set up a scenario for controlled radar detection and flyout. When CHASER was within 200m of the target, the radar would turn off, and probabilistic search would commence, with the camera off such that the search would continue indefinitely. As the search progressed and the belief map evolved, we recorded the estimated target probability captured within the field of view of the camera -- this is referred to as the Belief Gain, as it is our instantaneous likelihood of capturing the target in the FOV according to the target dynamics model. The results, in Figure \ref{fig:RHC_sitl}, show similar results from each of the different algorithm configurations. 

In this simple setup with no obstacles, we found that the similarity between max-belief and RHC algorithm indicated that RHC could perform at least as well as max-belief, making it a viable algorithm to use for field testing. Under more complex conditions, in which barriers or holes in the probability field might indicate buildings or other obstacles, or an improved target dynamics model offers more realistic updates to the belief map, we hypothesize that there may be additional benefit to using RHC, as it may be better able to explore non-spherical probability patterns. However, testing this hypothesis was beyond the scope of this project, and was left for future work.

Once the belief map and RHC algorithms were proved out in the SITL, we then tested them at field tests on Aug 15-16, 2022. To simulate a need for prolonged search using a path planning algorithm, we allowed target to fly in the field of view of the radar such that tracks were recorded. CHASER would then take off and attempt to observe the target, as we shut down the radar and flew target away from its detected position. 

In these tests, we used an assumption of maximum target speed of 10 m/s horizontally and 5 m/s vertically. We used a target belief map grid spacing of 20m and our grid spanned 400m in each cardinal direction with respect to the radar, vertically up 200m, and was bounded downward at the radar's altitude, to prevent waypoints from being generated below ground level. 

\begin{figure*}
\centering
\includegraphics[width=6in]{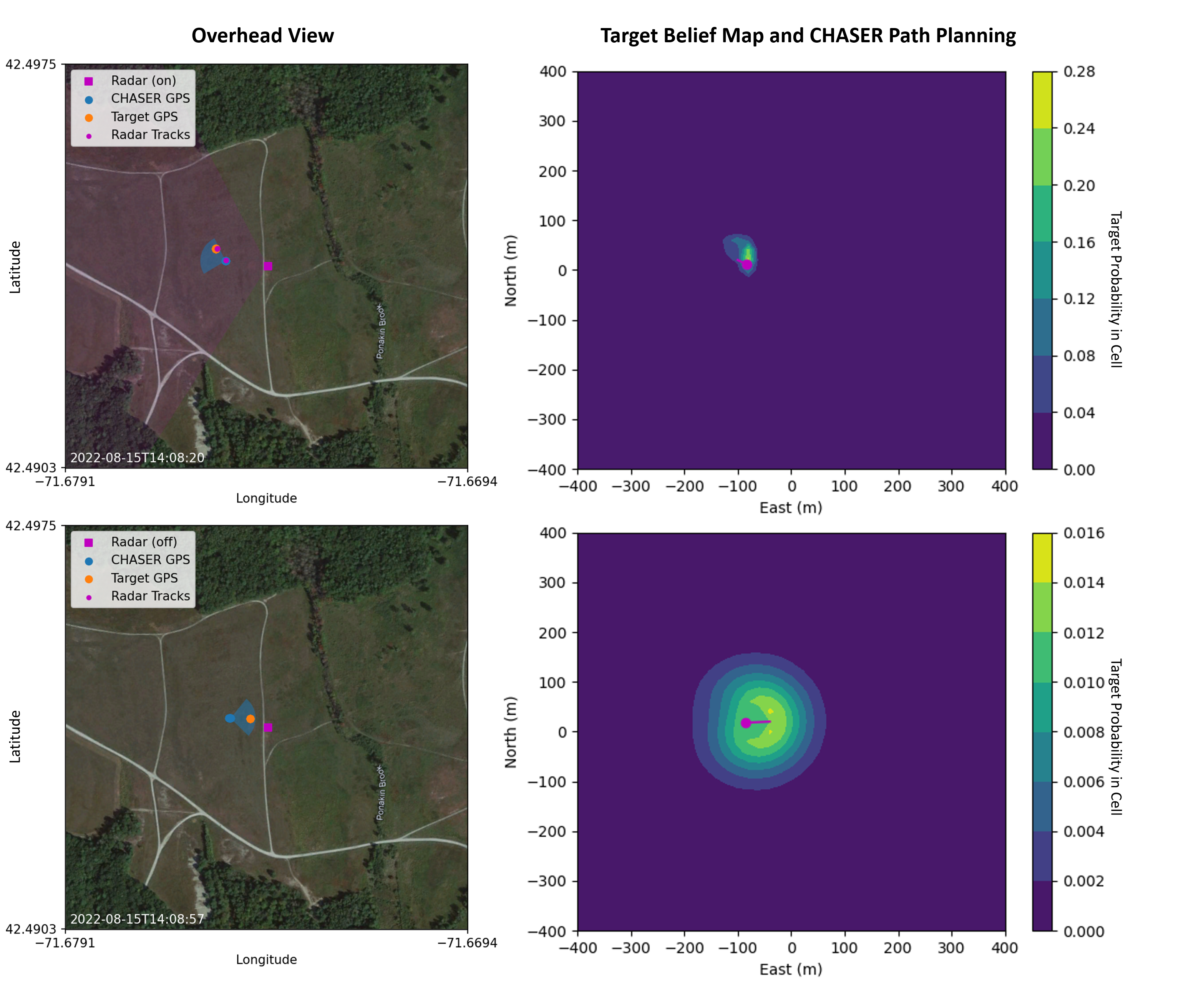}
\caption{\bf{15 Aug 2022 Field Test: (Top) Radar is producing tracks on both target and CHASER (left), and the target belief map (right) is collapsed to the immediate area surrounding target. Transition to CV-following has been disabled. (Bottom, 37 seconds later) The radar has been turned off, and target's position has become uncertain, growing according to the target dynamics model but adjusted with information from the sensor model. CHASER is searching for target using waypoints obtained from the Receding Horizon Controller. Note: target probabilities from the belief model are summed vertically for overhead map.}}
\label{Fig:BeliefMap}
\end{figure*}

In our first test, we disabled the transition to computer vision-based follow control, requiring that the search continued even if the target was detected by CHASER's camera. After initial detection of target tracks using the radar, the radar was shut off and the uncertainty in target position began to grow (See Figure \ref{Fig:BeliefMap}). CHASER's path was planned by the Receding Horizon Controller, which sent it on a course that took CHASER first west, then east, north, and west again, as it made two passes through the region where target was most likely to be. After these passes, it exited the transmission range of our ground equipment, and a manual override was initiated to retake control.


A second field test allowed us to test the transition to computer vision-based follow control. After an initial radar detection on the airborne target, we initiated an autonomous takeoff and flyout to the position of the detected target. 15.48 seconds after the radar detection, CHASER was airborne and confirmed its takeoff at an altitude of 20m AGL. From there, it began flying to the target using the RHC controller. After just 4.95 seconds, the target was acquired in the field-of-view of the camera, and the control mode was changed to computer vision-based PID follow control. This was prior to any additional waypoints were generated by the RHC path planner.

\section{Discussion}
\label{sec:discussion}
As demonstrated by the multitude of interacting components described in this paper, developing autonomous, reactive behaviors for sUAS is a fairly complex endeavor. Many lessons were learned in the course of two years of development and field testing of the autonomous sUAS testbed. Of primary importance, having a simulation environment in which to test algorithms prior to live flight was critical for debugging and evaluation. Other observations include:
\begin{enumerate}
    \item mounting the camera on a tilt-stabilizing gimbal is critical to mitigate against winds aloft and platform motion, while still minimizing coordinate transform complexity and error accumulation
    \item on-board GPU processing is critical; attempting to stream video to a ground-based processor introduced significant range limitations from the wifi antenna range limit, as well as significant lag.
    \item initial field testing did not sufficiently stress the RHC algorithm to demonstrate greater utility than a simpler max-belief algorithm; more complex testing is needed to explore differences in the algorithms
    \item some form of robust radar calibration is critical, whether it comes from better siting instrumentation or an on-the-fly calibration as discussed in Appendix \ref{appendix:RadarCalibration}
\end{enumerate}

\subsection{Boosting Small Teams}
In early stages of this work, as we designed an architecture to address the sUAS-chase mission described in Figure \ref{fig:chaseOV1}, we found the landscape of available sensor-guided autonomous sUAS capabilities to be limited. While commercial sUAS had robust built-in flight modes for limited autonomy, like RTL, waypoint navigation, and "Follow-me", their closed interfaces and inflexibility would make them useful only under constrained circumstances. As such, we found that a parallel set of tools with open interfaces and sensor inputs could provide a foundation for rapid iteration and deployment for complex small-team missions like the sUAS-chase.

A modular architecture for autonomous guidance controls was introduced with CHASER, with multiple components/tools implemented and tested in the field. In this paper, we've described our implementation and the performance of these individual components, including ground-based radar track correlation, an intelligent waypoint generation path planner for computer vision search, and a PID controller for following relative coordinates. However, to demonstrate the rapidly deployable nature of the architecture, we also configured the tools to address the sUAS-chase mission.

We found that in the process of configuring tools to the sUAS-chase mission, it was challenging to keep components from becoming coupled, and maintaining a federated architecture. To address this, we separated our mission-specific code from core mission utilities, allowing utilities to be "overridden" when tailored to a specific mission. Similarly, the configuration of these modules can be controlled with the use of mission-specific states, as noted in Section \ref{sec:arch}. However, we note that to address new missions in this way, some mission-specific code must be written and states defined. We find this to be a necessary and acceptable trade-off between the development of extensible building blocks of autonomy and the rapid adaptation to specific missions.

In the process of developing the core capabilities useful across multiple missions, we performed analysis to show which types of capabilities would enable the most small-team homeland protection-focused missions. We found that the most broadly applicable capabilities would be following global coordinates, taking off and flying out to global coordinates, and autonomous aerial CV search, so we addressed these with our work. Other high priority capabilities that were developed but not discussed extensively in this paper include ownship safety checks, ingestion of GPS-based cues for following, and streaming live video back to an operator. Additional undeveloped capabilities that would meet the needs of multiple future missions include infrared computer vision object detection and stereo vision-based ranging. We aim to develop these capabilities in future work.




\subsection{Continued Improvement of Modular Autonomy}
While we have developed the baseline functionality for many of the components required for modular autonomy, additional work is required in some cases to improve their robustness to unexpected conditions. As noted in our results (Section \ref{sec:testing}), we relied on a GPS on the target platform to provide us with accurate estimates of target range during tracking for the sUAS-chase mission. Despite efforts to correlate bounding box size during object detection with target range, such estimates proved too variable to use with the relative position controller, even with strong smoothing. Our intention was to replace the bounding box-based method with stereoscopic camera range estimation \cite{brunet2020stereo}, but due to on-board computational constraints from hardware limitations, we were unable to implement such an algorithm. We continue to believe that stereo-based ranging would be a good option for reducing variance in range estimation, and hope to implement this capability in the future.

We also noted in the results that we observed similar belief gain performance in the SITL between the Receding Horizon Controller and a "Max Belief" waypoint generation path planner (Section \ref{sec:testing}, Receding Horizon Controller). Our belief model, or target probability map, has an evolution over time largely driven by the target dynamics model. Path planning is predicated on the accuracy of the belief model, and therefore the target dynamics model. For this initial implementation, we chose a simple target dynamics model that represents the state transitions of the target using a 3D convolution over the current belief model. We believe that improving on this target dynamics model by incorporating the target velocity, known scene features, or other available information, would improve the belief model and lead to more intelligent path planning using the RHC algorithm. A more realistic target dynamics model could mean that the computer vision search volume is made smaller, enabling better performance in acquiring the target during search.

Lastly, despite our focus on developing a modular set of capabilities, our experience configuring the modules to execute a mission is limited to our testing of the sUAS-chase mission. We believe that additional lessons would be learned by reconfiguring the modules to address a new mission, like search and rescue assistance or remote emplacement. Such a task would help to further segregate mission-specific functionality from general utilities, and continue to improve the modularity of the CHASER system.

\section{Conclusion}
\label{sec:conclusion}


After assessing the landscape of commercial sUAS autonomous functionality, we decided to build out a set of sensor-guided AI-enabled controls tools that would lower barriers to sUAS adoption on small teams--specifically those with homeland protection-focused missions. In this paper, we have introduced many of the components developed for our CHASER testbed, describing their implementation and performance. We paid particular attention to their integration and configuration in a specific scenario, the sUAS-chase mission, and demonstrated the modularity of the developed tools through the autonomous completion of this mission. Beyond a description of the capabilities in the testbed and their performance, we also consider their utility to small teams carrying out additional missions, and propose the development of additional capabilities to supplement the testbed. We envision agencies and teams leveraging the capabilities introduced in this work for the rapid development and deployment of autonomous sUAS to meet present and future homeland protection mission needs.

\acknowledgments
The authors thank the MIT Lincoln Laboratory Homeland Protection Line and Allocated fund for funding this project.

\bibliographystyle{IEEEtran}
\bibliography{references.bib}
\thebiography

\begin{biographywithpic}{Keegan Quigley}{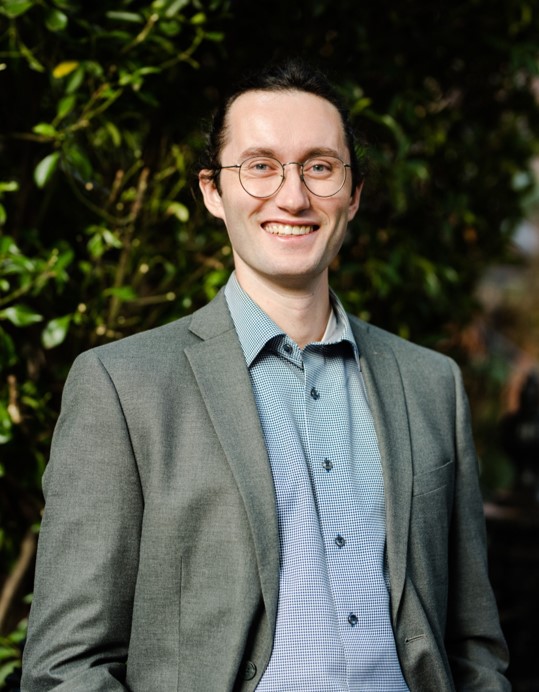}
is an associate member of the technical staff at MIT Lincoln Laboratory. He received his Sc.B. in Engineering Physics from Brown University, conducting research in remote sensing and photonics. At Lincoln Laboratory, he has worked on systems analysis and prototyping of small Uncrewed Aerial Systems (sUAS) and counter sUAS technologies, integrating physical system components, sensors, and algorithms to develop new capabilities. Mr. Quigley is now a member of the Artificial Intelligence Technology group at the laboratory, where his research focuses on deep learning algorithms and computer vision in both the autonomous vehicle and medical diagnostics domains.
\end{biographywithpic}

\begin{biographywithpic}
{Luis Alvarez}{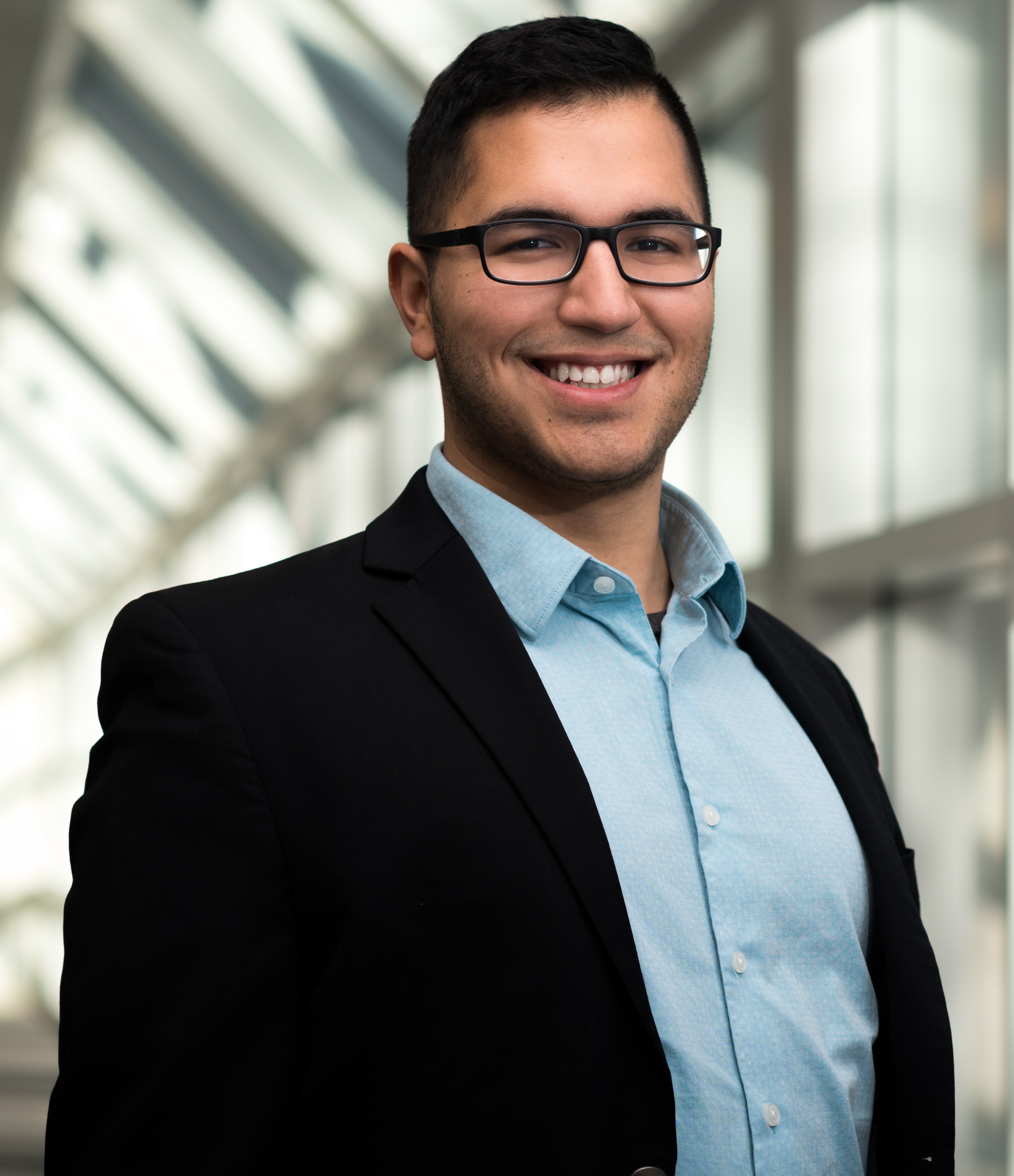}
received his B.S. degree in Aerospace Engineering in 2013 from University of Florida and a S.M. in Aeronautics and Astronautics from the Massachusetts Institute of Technology in 2016. He is currently a Technical Staff at MIT Lincoln Laboratory within the Homeland Protection and Air Traffic Control division. His current research is focused on applied machine learning for Advanced Air Mobility (AAM), strategic planning for Unmanned Aerial Systems (UAS) traffic management, and counter-UAS autonomous capabilities. He leads the development of the decision-making logic within ACAS X, a family of collision avoidance systems using Dynamic Programming and Markov Process. He also co-leads the development of the artificial intelligence testbed for AAM in which the team is rapidly adapting and testing machine learning algorithms for air traffic management. He is an active member of the AIAA Aviation Forum, Air Transportation Systems technical committee.

\end{biographywithpic}

\begin{biographywithpic}
{Virginia Goodwin}{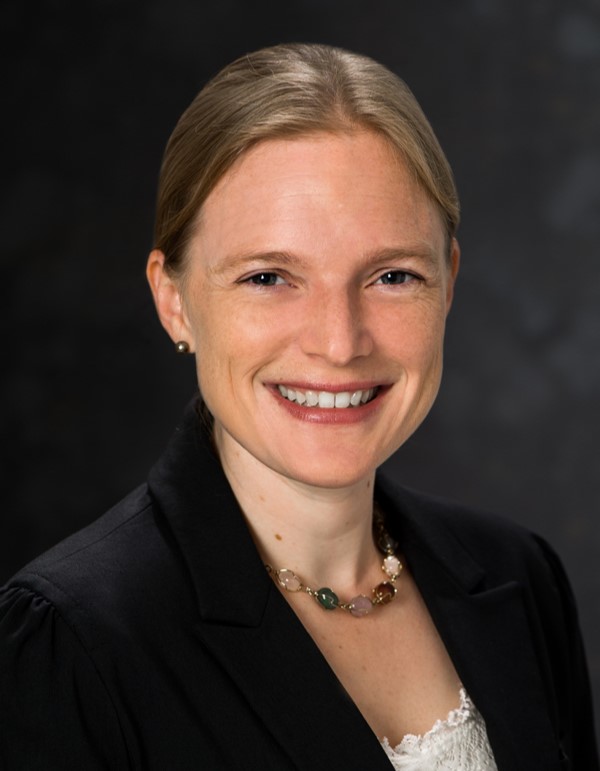}
is a member of the technical staff at MIT Lincoln Laboratory in Lexington, Massachusetts.  Ms. Goodwin received a B.A. degree in Physics from Wellesley College, and a M.S. degree in Engineering, with a focus on Information Theory, from Harvard University. Ms. Goodwin has worked in the fields of ballistic missile defense, homeland defense and air traffic control.  Her current work includes systems analysis for counter small uncrewed aerial systems (C-sUAS) defense technology, developing technology for sUAS autonomy, and machine learning algorithms for air defense. Her research interests include statistical pattern recognition, decision support systems, and the ethical employment of machine learning systems.
\end{biographywithpic}

\begin{biographywithpic}
{Justin Yao}{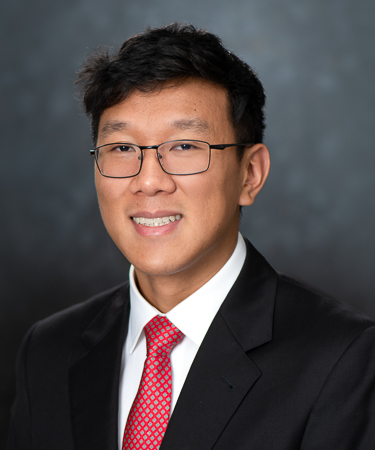}
is an associate member of technical staff at MIT Lincoln Laboratory. He received his B.S. in Computer Science and M.S. in Computer Science from the Georgia Institute of Technology, focusing on artificial intelligence and machine learning. At Lincoln Laboratory, he has worked on prototyping small Uncrewed Aerial Systems (sUAS) and counter sUAS technologies and developing computer vision algorithms and machine learning systems for homeland protection and air defense. His research interests include developing deep learning algorithms for advanced optical sensing and autonomous systems.

\end{biographywithpic}

\begin{biographywithpic}
{Yousef Salaman Maclara}{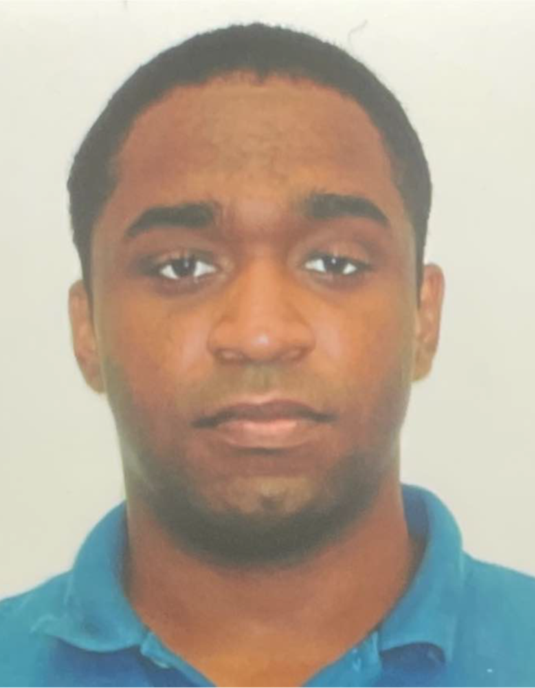}
is a Robotics Ph.D. student at Oregon State University. He received his B.S. in Electrical Engineering in the University of Puerto Rico, Mayag\"uez, with a focus in control systems. His research interests lie in using properties that arise in dynamical systems to design control algorithms.

\end{biographywithpic}

\newpage
\appendices{}              

\section{Radar Calibration}  \label{appendix:RadarCalibration}
\subsection{Methods}
To use the sUAS radar to establish tracks on target, a method for calibration of the radar's orientation and position was required. The use of sUAS and reflective objects for precise calibration of weather radars has been explored \cite{UAVRadarCal}\cite{WeatherRadarCal}, but we aimed to develop a method for calibration that could be performed in under 10 minutes with the Echodyne Echoguard radar, without the use of dedicated GPS hardware on the radar. Our method is most similar to that employed by \textit{Nowak \& Maksimiuk, 2019} \cite{Nowak}.

To calibrate, take a best guess of its position and orientation using a handheld GPS and iPhone compass. We would then fly one target sUAS into the field-of-view of the radar and record the track data outputs, along with the GPS coordinates of the sUAS. The first stage of the algorithm aimed to select a singular radar track from this set to use as the calibration track. As a simple first pass, the track that lasted the longest was selected as the candidate track. In the second stage of the algorithm, we used a least squares fit to find the radar orientation that minimized the difference between GPS position and Track positions in global coordinates, with measurements from each sensor binned into 0.5 second windows and averaged. This orientation was then substituted for the initial guess for subsequent testing.

\subsection{Testing}
Before using the radar for field testing, we ran our calibration routine to determine orientation parameters (Yaw, Pitch, Roll). Latitude, longitude, and altitude were measured using a handheld GPS device. In our July 2022 testing, our calibration flight lasted 3 minutes, 36 seconds, and we recorded 22 tracks with track lifetimes ranging from 0.5 to 29.9\% of the flight. Of these, the longest was selected for least squares fitting, and the fit resulted in residual ENU errors (one standard deviation) of 0.72, 0.29, and 1.89 meters with respect to the windowed GPS positions. In all of our flight tests from 2022, we were able to identify a track that could be used for fitting, with residuals no more than 2.5 meters from the GPS track. We note that during this calibration flight, the pilot flew circles and vertical passes in front of the radar, leading to a diverse set of points for calibration. However, the fit yielded larger vertical residuals than horizontal residuals, indicating that more diversity of elevation points may have been needed.

During post-processing, we were able to test an extension of our algorithm, where all 22 tracks were assessed using a least squares fit, and the track producing the lowest residuals after fitting was selected. This method was found to work better than the original method of selecting the longest track for fitting. The best track had one standard deviation errors in ENU of 0.63, 0.50, and 1.09 with respect to the windowed GPS measurements. 

A further extension was explored for calibration using least squares fitting of orientation \textit{and} position parameters (latitude, longitude, altitude). It was found that when fitting these six parameters, one-SD residuals routinely exceeded 100 meters, indicating poor fitting. We hypothesize that longer tracks and initialization of fitting parameters with known values may lead to better performance for six-parameter fitting. For the remainder of flight tests in this study, we used only the original algorithm. 

\clearpage
\setboolean{@twoside}{false}



\section*{Supplementary material}

# Model Card: CHASER inc4

## Model Details

Developed by: Justin Yao, Virginia Goodwin, Luis Alvarez, Alex Showalter-Bucher
Training Date: 06/01/2021
Version: 1
Architecture: YOLOv2 Convolutional Neural Network
References:
  - YOLOv2 paper: https://arxiv.org/abs/1612.08242
  - Darknet GitHub Repo: https://github.com/AlexeyAB/darknet/
Citation Details: None
License: None
Contact Information: justin.yao@ll.mit.edu or virginia.goodwin@ll.mit.edu

- Configuration and Training Parameters:
  - Uses `darknet19_448.conv.23` as pre-trained weights
  - Other model architecture and parameters do not differ from the standard YOLOv2 architecture except that final output layers have been adjusted to support the new label set

Learning Schedule:

| Parameter | Value |
|---|---|
| Batch Size | 256 |
| Number of Mini Batches | 16 |
| Burn-in Period | 1,000 |
| Total Batches | 500,200 |
| Learning Rate | 0.001 |
| Learning Rate Decay Policy | Step-wise<br>- Decayed by .1 at 100,000, 200,000, 300,000, and 400,000 |
| Optimizer | Nesterov Accelerated Gradient Descent<br>- Momentum = 0.9<br>- Decay = 0.0005 |

Data Augmentation:
- Network Input Resolution Augmentation:
  - Default Input Resolution: 416 × 416 × 3
  - Default input height & width randomly multiplied/divided by [1.0, 1.4]
- Per-Image Augmentation:
  - Height & width randomly cropped/scaled by [.4, 1.6]
  - Color Augmentation:
    H – randomly rotated by [-36, 36] degrees
    S – randomly multiplied/divided by [1.0, 1.5]
    V – randomly multiplied/divided by [1.0, 1.5]
  - Bilateral filter applied randomly 50% of the time
- Multi-Image Augmentation:
  - Mosaic
  - MixUp

Label Set:

| Class ID | 0 | 1 | 2 | 3 | 4 |
|---|---|---|---|---|---|
| Class Name | sUAS | Airplane | Bird | Car | Person |

## Intended Use

- Primary Intended Uses:
  - Detecting S1000 drones against bright sunny grassy fields with forest surroundings using color images
- Primary Intended Users:
  - MIT LL developers
- Out-of-scope Use Cases:
  - Detecting S1000 drones in environments other than bright sunny grassy fields
  - Detecting objects besides S1000 drones, including other drone types
  - Running detection on other imaging types/spectra (e.g., grayscale, thermal IR)

## Performance Factors

Relevant Performance Factors:

Environment:
- The performance of object detectors in general is known to be impacted by image background, such as colors and textures in the surrounding environment
- The performance of the model has been observed to change under different lighting conditions where the appearance of the drone can change depending on the amount of light bouncing off of its reflective surfaces. Additionally, the current model has not been trained to detect drones in low light settings.

Cameras:
- The performance of object detectors is known to be impacted by a camera's hardware and software, including lens, ISO, image stabilization, automatic gain control, etc.

Evaluated Performance Factors:

Environment:

Ft. Devens
Sunny grassy field with gravel roads, small water stream, and forest surroundings; recordings held in early spring and summer

Gardner Airport
Sunny grassy field with sandy patches, paved and gravel roads, tarmac surrounded by chain linked fence, and forest surroundings; recordings held in mid-summer

Cameras:
Multiple cameras were used to capture video for training and evaluation. GoPros were mounted on the frame of the platform for initial data collections and as back-up to the e-con. The e-con was the primary camera for real-time processing

GoPro Hero 4 Black
- FOV[1]: 122.64°(H), 72.22°(V), 138.48°(D)
- Focus: fixed
- Recording format: H.264
- Resolution: 1920 × 1080
- Frame rate: 30Hz

e-con 4K e-CAM130CUXVR: fixed focus
- FOV: 95°(H), 73°(V), 134°(D)
- Focus: fixed
- Recording format: H.264
- Resolution: 1920 × 1080
- Frame rate: 30Hz

## Metrics

Model Performance Measures:

| Term | Definition |
| --- | --- |
| Average Precision (AP[2]) | average value of the precision-recall curve at a given IoU threshold for varying confidence threshold values; used to generally measure detection performance |
| Average Intersection over Union (IoU) | average intersection area divided by union area; used as a measure of how well bounding boxes line up with ground truth |
| Inference Speed[3] | the FPS at which the model can process images |
| @ Confidence Threshold | minimum confidence score for predicting a detection |
| Precision | measure of what percentage of predicted detections are true detections |
| Recall | measure of what percentage of ground truth has a predicted detection; also known as probability of detection |
| $F_1$ Score | harmonic mean of the precision and recall; used as an aggregate score of precision and recall |

- Decision Thresholds:
  - 25% confidence threshold during validation and testing
  - 20% during live runs
    Selected empirically based off performance during field tests
- Approaches:
  - Metrics were evaluated on a held-out validation set and final field test dataset
  - Performance was also qualitatively evaluated during field tests

[1]This FOV was inferred from documentation   [2]We utilize 11-point interpolated AP (PascalVOC AP)   [3]As measured on the NVIDIA Jetson AGX Xavier

## Evaluation Data

- Datasets:
  - Validation data is listed in `chaser_inc4_val.txt`[i]
  - Final flight test data listed in `chaser_inc4_test.txt`[i] (5 minutes and 27 seconds of flight time collected at Gardner Airport, 16 July 2021)
- Motivation:
  - Validation and test dataset contain videos recorded at Ft. Devens and Gardner Airport that include top and side views of the S1000 drone in the air and on the ground, which closely matches our use case
- Preprocessing:
  - Images have pixel values normalized to [0, 1] and have resolution rescaled to 832 × 832 (primary setting) or 416 × 416 (additionally tested setting) before inference

## Training Data

- Datasets:
  - All images used for training are listed in `chaser_inc4_train.txt`[i]
- Motivation:
  - Training dataset contains videos recorded at Ft. Devens that include top and side views of the S1000 drone in the air and on the ground, which closely matches our use case
- Preprocessing:
  - Dataset was compressed from PNG to JPEG to enable easier handling. No other preprocessing was performed prior to training with Darknet

## Quantitative Analysis

### Gardner Airport Final Test Flight (`chaser_inc4_test.txt`[i]) Metrics

| Input Resolution | sUAS AP@0.50 | Average IoU | Inference Speed | @0.25 Confidence Threshold | | |
|---|---|---|---|---|---|---|
| | | | | Precision | Recall | F$_1$ Score |
| 416 × 416 | 9.49% | 0.1830 | 19 – 24 | 0.29 | 0.15 | 0.20 |
| 832 × 832 | 12.37% | 0.1484 | 8 – 10 | 0.25 | 0.21 | 0.23 |

### Validation Set (`chaser_inc4_val.txt`[i]) Metrics

| Input Resolution | sUAS AP@0.50 | Average IoU | Inference Speed | @0.25 Confidence Threshold | | |
|---|---|---|---|---|---|---|
| | | | | Precision | Recall | F$_1$ Score |
| 416 × 416 | 71.92% | 0.6658 | 19 – 24 | 0.86 | 0.71 | 0.78 |
| 832 × 832 | 82.33% | 0.6590 | 8 – 10 | 0.84 | 0.82 | 0.83 |

- Other Observations:
  - Running inference at 832 × 832 resolution especially improves false negative rate compared to running at 416 × 416 resolution

## Caveats and Recommendations

- Model was observed to not do well on detecting drones over sandy patches in the grass
- We recommend further data collection and training to improve the model performance
  - We recommend increasing diversity in the dataset as we believe the model is overfit to sunny grassy fields
- We recommend switching to a more up to date model architecture, such as YOLOv4
- We recommend using a deep learning inference optimizer and runtime such as TensorRT to improve model FPS
- The object detection model was designed for the potential to detect 5 categorical classes (sUAS, airplane, bird, car, person) but currently only has the capability to detect sUAS. This is largely because the current training, validation, and testing datasets only have sUAS labeled; all instances of all other classes are unlabeled. It would be ideal if these other class instances were annotated so that the model would be able to detect its full range of classes. If this is not a feasible option, then the other four classes should be ignored or removed

## Ethical Considerations

**Data**: No sensitive data was used during model development

**Human Life**: Model is not intended to inform decisions about human centric matters

**Mitigations**: No ethical risk mitigation strategies were applied during model development

**Risks and Harms**: No known ethical risks in model usage

**Use Cases**: No known use cases of particular ethical concern

[i]These are specific splits of the CHASER sUAS dataset. Please contact Justin or Virginia for access.

# C. DATASHEET

**CHASER sUAS Data**

### Motivation

**Why was this dataset created?**
The dataset was created as there was a lack of open-source drone video especially for the S1000 drone. The specific task in mind for this dataset was to train an object detection model to detect the S1000 drone while the camera is in flight on another drone.

**Who created this dataset?**
MIT LL CHASER team.

**Who funded the creation of this dataset?**
MIT LL Homeland Protection Allocated Funding.

**Any other comments?**
None.

### Composition

**What is the dataset comprised of?**
The dataset is composed of EO videos captured from a variety of cameras and primarily collected from Ft. Devens, MA and Gardner Airport, MA. The videos feature the S1000 drones primarily against sunny grassy fields. Only bounding box track annotations of the S1000 are present.

**How many instances are there in total?**
The dataset contains 267,019 images with the following label breakdown:

| Object | # of Instances | # of Images |
|---|---|---|
| S1000 sUAS | 101,264 | 101,264 |
| Total | 101,264 | 267,134 |

**Does the dataset contain all possible instances or is it a sample of instances from a larger set?**
The data are only representative of a very narrow segment of drone imagery in limited background/lighting conditions. Primarily grassy (brown & green) backgrounds with bright, outdoor lighting.

**What data does each instance consist of?**
Each instance in the dataset is an unprocessed and uncompressed PNG image file that corresponds to a video frame.

**Is there a label or target associated with each instance?**
Only bounding box track annotations of the S1000 are annotated if present in the image.

**Is information missing from individual instances?**
No.

**Are relationships between individual instances made explicit?**
Images from the same video are grouped together in sequence. For annotations, tracks maintain continuity for bounding boxes on adjacent frames. However, each track of the sUAS is separate from the others (e.g., one track becomes two tracks if the sUAS becomes fully occluded), even though during the flight tests we know that it is the same sUAS in every track.

**Are there recommended data splits?**
We used chaser_inc5_train.txt for training and chaser_inc5_val.txt for validation splits. Summary descriptions of these splits can be found in chaser_inc5_train_stats.txt and chaser_inc5_val_stats.txt, respectively. We recommend splitting the dataset into contiguous chunks to prevent too many adjacent video frames from being shared in training, validation, and testing splits.

**Are there errors, noise, redundancies?**
Yes. There exists human error and noise in bounding box placement. We made best efforts to keep error and noise at a minimum. For redundancies, adjacent frames collected from the camera at 60 Hz or 30 Hz may be highly correlated and somewhat redundant.

**Is the dataset self-contained, or does it link to or otherwise rely on external resources?**

**CHASER sUAS Data**

The dataset is self-contained.

**Does the dataset contain data that might be considered confidential?**

No. However, this data was collected with MIT LL equipment and has not been approved for public release. It is currently only available within the MIT LL community (10/5/22).

**Does the dataset contain data that, if viewed directly, might be offensive, insulting, threatening, or might otherwise cause anxiety?**

No.

**Does the data relate to humans?**

No, except humans occasionally appear in the video (not labeled).

**Is it possible to identify individuals, either directly or indirectly from the dataset?**

Members of the MIT LL CHASER team are present and identifiable in the dataset.

**Any other comments?**

None.

## Collection

**How was the data associated with each instance acquired?**

Planned flight tests.

**What mechanisms or procedures were used to collect the data?**

Video was collected using EO cameras (GoPro Hero 4 Black and e-con 4K e-CAM130CUXVR cameras) and the S1000 drone. Data was manually curated and then uploaded into CVAT for annotation.

**Is data a subset, if so what was sampling strategy?**

The data represent as much imagery as we were able to collect and label.

**Who was involved in the data collection process and how were they compensated?**

Data was entirely collected and annotated by the MIT LL CHASER team.

**Over what timeframe was data collected?**

Periodically December 2020 – August 2021.

**Were there any ethical review boards?**

No.

**Were the individuals in the dataset notified about the data collection?**

Yes.

**Did the individuals in the dataset consent to the collection and use of their data?**

Yes.

**If consent was obtained, were the consenting individuals provided with a mechanism to revoke their consent in the future or for certain uses?**

No.

**Has an analysis of the potential impact of the dataset and its use on data subjects been conducted?**

Impact to the individuals in the dataset should be minimal.

**Any other comments?**

None.

## Preprocessing

**Was any preprocessing/cleaning/labeling of the data done?**

No. However, the occlusion flag was used during annotation, so it is possible to filter out occluded bounding boxes in the current dataset if desired.

**Was the raw data saved?**

Yes.

**Is the software that was used to preprocess/clean/label the data available?**

We used CVAT to label our data. The source code and documentation are available at: https://github.com/opencv/cvat

**CHASER sUAS Data**

**Any other comments?**
None.

## Uses

**Has the dataset been used already?**
The dataset is currently being used to train object detection models for detecting the S1000 drone.

**Is there a repo to the code that uses this data?**
Our code repository is not currently approved for public release.

**What (other) tasks could the dataset be used for?**
As it currently is, the dataset is best used for developing algorithms for detecting sUAS.

**Is there any aspect of the composition/collection/ preprocessing that may impact future use?**
We did not label other objects in the video, including people, cars, tents, tables, small aircraft, etc. Additionally, the dataset consists mostly of video of the S1000 drone in sunny grassy field settings, so there is not too much diversity or variability in the dataset.

**Are there any tasks for which this dataset should explicitly NOT be used?**
No.

**Any other comments?**
None.

## Distribution

**Will/Can this data be distributed outside of MIT LL?**
MIT LL internal: Yes. External: TBD.

**How will the dataset be distributed internally at MIT LL?**
Please contact Virginia or Justin for access to the CVAT server.

**Do any export controls or other regulatory restrictions apply to the dataset or to individual instances?**
No.

**Any other comments?**
None.

## Maintenance

**Who is supporting/hosting/maintaining the dataset?**
Virginia Goodwin, Justin Yao, Michelle Tiner at MIT LL.

**How can the owner/curator/manager of the dataset be contacted?**
virginia.goodwin@ll.mit.edu or justin.yao@ll.mit.edu

**Is there an erratum?**
No.

**Will the dataset be updated?**
If/when we label additional data we will update the dataset.

**Are there applicable limits on the retention of the data associated with the instances?**
No.

**Will older versions of the dataset continue to be supported/hosted/maintained?**
Yes, we will maintain copies of older versions of the dataset.

**If others want to extend/augment/build on/contribute to the dataset, is there a mechanism for them to do so?**
Please contact Virginia or Justin if you would like to contribute to the dataset.

**Any other comments?**
None.

## Additional Information
None.



\end{document}